%% file: main.tex
\documentclass[accepted]{uai2025} % after acceptance, for a revised version; 
\title{The Relativity of Causal Knowledge}
\author[1,2]{\href{mailto:<gabriele.dacunto@uniroma1.it>?Subject=About "The Relativity of Causal Knowledge"}{Gabriele D'Acunto}{}}
\author[3]{\href{mailto:<cbattiloro@hsph.harvard.edu>?Subject=About "The Relativity of Causal Knowledge"}{Claudio Battiloro}{}}
\affil[1]{%
    Information Engineering, Electronics and Telecommunications Dept.\\
    Sapienza University\\
    Rome, Italy
}
\affil[2]{%
    National Inter-University Consortium for Telecommunications (CNIT)\\
    Parma, Italy
}
\affil[3]{%
    Biostatistics Dept.\\
    Harvard University\\
    Cambridge, MA, USA
}

\input{macros}

\begin{document}

\maketitle

\begin{abstract}
Recent advances in \emph{artificial intelligence} reveal the limits of purely predictive systems and call for a shift toward causal \emph{and} collaborative reasoning.
%% IDEA
Drawing inspiration from the revolution of Grothendieck in mathematics, 
we introduce the \emph{relativity of causal knowledge}, which posits structural causal models (SCMs) are inherently imperfect, subjective representations embedded within networks of relationships.
%% CONTRIBUTION
By leveraging category theory, we arrange SCMs into a functor category and show that their observational and interventional probability measures naturally form convex structures.
This result allows us to encode non-intervened SCMs with convex spaces of probability measures.
Next, using sheaf theory, we construct the \emph{network sheaf and cosheaf of causal knowledge}. 
These structures enable the transfer of causal knowledge across the network while incorporating interventional consistency and the perspective of the subjects, ultimately leading to the formal, mathematical definition of \emph{relative causal knowledge}. 
\end{abstract}

\input{sections/intro}
\input{sections/scm_cs}
\input{sections/causal_knowledge}
\input{sections/cellular_sheaf}
\input{sections/discussion}
\input{sections/conclusion}
%\clearpage

\begin{contributions} % will be removed in pdf for initial submission 
					  % (without ‘accepted’ option in \documentclass)
                      % so you can already fill it to test with the
                      % ‘accepted’ class option
    The Authors equally contributed to the original idea, the thorough conceptual development of the framework, and the paper writing.
    GD developed the category-theoretic framework, and established and proved the related theoretical results.
    GD and CB codeveloped the sheaf-theoretic characterization of RCK.
\end{contributions}

\begin{acknowledgements} % will be removed in pdf for initial submission,
						 % (without ‘accepted’ option in \documentclass)
                         % so you can already fill it to test with the
                         % ‘accepted’ class option
The work of Gabriele D'Acunto was supported by \emph{(i)} the European Union under the Italian National Recovery and Resilience Plan (NRRP) of NextGenerationEU, partnership on `` Telecommunications of the Future'' (PE00000001 - program `` RESTART''), 
and \emph{(ii)} the SNS JU project 6G-GOALS \cite{strinati2024goal} under the EU's Horizon program Grant Agreement No 101139232.

The work of Claudio Battiloro was supported by the National Institutes of Health (NIH) Grant 1R01ES037156-01.

The Authors thank Dr. Hans Riess from Duke University and Dr. Fabio Massimo Zennaro from University of Bergen for valuable comments on a preliminary version of the work.
\end{acknowledgements}

\bibliographystyle{plainnat}
\renewcommand{\bibsection}{\subsubsection*{References}}
\bibliography{bibliography}

\newpage

\onecolumn

\title{The Relativity of Causal Knowledge\\(Supplementary Material)}
\maketitle

\appendix
\input{appendix/background}
\input{appendix/examples}

\input{appendix/invertibility}
\input{appendix/proofs}
\input{appendix/disc_ex}

\end{document}

%% file: macros.tex
\usepackage[english]{babel}

\usepackage{enumitem}
\usepackage{type1cm} % type1 computer modern font
\usepackage{graphicx} % advanced figures
\usepackage{xspace} % fix space in macros
\usepackage{balance} % to better equalize the last page
\usepackage{booktabs} % nicer tables
\usepackage{multirow} % multi rows for tables
\usepackage{fontawesome5}
\usepackage[font={bf}, tableposition=top]{caption} % captions on top for tables
\usepackage{subcaption} % subfloats
\usepackage{bold-extra} % bold + {small capital, italic}
\usepackage[vlined,linesnumbered,ruled,noend]{algorithm2e} % algorithms
\usepackage{microtype} % compress text
\usepackage{siunitx} % \num for decimal grouping
\usepackage{xfrac} % nicer slanted fractions
\usepackage{amsmath}
\usepackage{amssymb}
\usepackage{mathtools}
\usepackage{amsfonts}
\usepackage{amsthm}
\usepackage{bbm}
\usepackage[square,numbers]{natbib} % better references
\usepackage{cleveref} % smart references
\usepackage[hyperpageref]{backref} % back references
\usepackage{hyphenat} % name in single line
\usepackage{csquotes}
\usepackage[show]{chato-notes} % notes
\usepackage{authblk}
\usepackage{bm}
\usepackage{xspace}
\usepackage{physics}
\usepackage{newunicodechar}
\newunicodechar{⟓}{\ensuremath{\uplus}}
\usepackage[dvipsnames]{xcolor}
\usepackage{tikz}
\usetikzlibrary{positioning}
\usetikzlibrary{bayesnet}
\usetikzlibrary{arrows}
\usetikzlibrary{shapes}
\usetikzlibrary{fit}
\usepackage{tikz-cd}
\usepackage{bbold}
\usetikzlibrary{calc, positioning, arrows.meta, shapes.geometric, fit, backgrounds}
\usepackage{pgf}
\usepackage{thmtools,thm-restate}
\usepackage{tikz-3dplot}
\usepackage{tcolorbox}

\newcommand{\spara}[1]{\smallskip\noindent\textbf{#1}}

\newenvironment{squishlist}
{\begin{list}{$\bullet$}
 {\setlength{\itemsep}{0pt}
     \setlength{\parsep}{1.5pt}
     \setlength{\topsep}{1.5pt}
     \setlength{\partopsep}{0pt}
     \setlength{\leftmargin}{1em}
     \setlength{\labelwidth}{1em}
     \setlength{\labelsep}{0.5em} } }
{\end{list}}

\renewcommand*\backref[1]{\ifx#1\relax \else (Cited on #1) \fi}

\hypersetup{
   bookmarks, pdftex,
   colorlinks=true,
   pagebackref=true, backref=page,
   linkcolor={red!50!black},
   filecolor={green!50!black},
   citecolor={green!50!black}, 
   urlcolor={green!40!black},
}

\graphicspath{{fig}}

\newtheoremstyle{mytheoremstyle}%
  {2pt}    % Space above
  {2pt}    % Space below
  {\itshape}  % Body font
  {}          % Indent amount
  {\bfseries} % Theorem head font
  {}          % Punctuation after theorem head
  { }         % Space after theorem head
  {}          % Theorem head specification

\theoremstyle{mytheoremstyle}

\newtheorem{definition}{Definition}

\newtheorem{remark}{Remark}
\newtheorem{example}{Example}

\newcommand{\reall}{\ensuremath{\mathbb{R}}\xspace}

\newcommand{\zeros}{\ensuremath{\boldsymbol{0}}\xspace}

\newcommand{\identity}{\ensuremath{\mathbf{I}}\xspace}
\newcommand{\catidentity}{\ensuremath{\mathrm{Id}}\xspace}
\newcommand{\parents}{\ensuremath{\mathcal{P}}\xspace}
\newcommand{\myendogenous}{\ensuremath{\mathcal{X}}\xspace}
\newcommand{\myexogenous}{\ensuremath{\mathcal{Z}}\xspace}
\newcommand{\myfunctional}{\ensuremath{\mathcal{F}}\xspace}
\newcommand{\mymixing}{\ensuremath{\mathcal{M}}\xspace}

\newcommand{\mixing}{\ensuremath{\mathbf{M}}\xspace}

\newcommand{\Vect}{\ensuremath{\mathsf{Vect}_{\reall}}\xspace}

\newcommand{\Eprod}[3]{\ensuremath{\langle #1, \, #2 \rangle_{#3}}\xspace}

\newcommand{\Hilb}{\ensuremath{\mathsf{Hilb}_{\reall}}\xspace}
\newcommand{\IPVect}{\ensuremath{\mathsf{preHilb}_{\reall}}\xspace}

\newcommand{\Poset}{\ensuremath{(\mathcal{P}, \leqslant)}\xspace}
\newcommand{\CSprob}{\ensuremath{\mathsf{CSprob}}\xspace}

\newcommand{\CK}[1]{\ensuremath{\mathsf{CK}\left(#1\right)}\xspace}
\newcommand{\faceincidenceposet}{\ensuremath{(\Pi,\trianglelefteq)}\xspace}

\newcommand{\X}{\ensuremath{\mathbf{X}}\xspace}

\newcommand{\Z}{\ensuremath{\mathbf{Z}}\xspace}

\newcommand{\scm}[1]{\ensuremath{\mathsf{M}^{#1}}\xspace}
\newcommand{\intervention}{\ensuremath{\mathcal{I}}\xspace}

\newcommand{\hard}{\ensuremath{\mathcal{H}}\xspace}
\newcommand{\soft}{\ensuremath{\mathcal{S}}\xspace}

\newcommand{\indicatorf}[2]{I_{#1}\left(#2\right)}

\newcommand{\blfootnote}[1]{%
  \begingroup
  \renewcommand\thefootnote{}\footnote{#1}%
  \addtocounter{footnote}{-1}%
  \endgroup
}

\newcommand{\myendogenousvals}{\ensuremath{\mathcal{V}}\xspace}
\newcommand{\myexogenousvals}{\ensuremath{\mathcal{U}}\xspace}
\newcommand{\Prob}{\ensuremath{\mathsf{Prob}}\xspace}

\newcommand{\I}{\ensuremath{\mathsf{I}}\xspace}
\newcommand{\ancestors}{\ensuremath{\mathcal{A}}\xspace}
\newcommand{\SCMcat}{\ensuremath{\mathsf{SCM}(\I,\Prob)}\xspace}

\newcommand{\NI}{\ensuremath{\mathsf{NI}(\I,\Prob)}\xspace}
\newcommand{\Ccat}{\ensuremath{\mathsf{C}}\xspace}
\newcommand{\Dcat}{\ensuremath{\mathsf{D}}\xspace}

\newcommand{\abst}{\ensuremath{\boldsymbol{\alpha}}\xspace}
\newcommand{\Rset}{\ensuremath{\mathcal{R}}\xspace}
\newcommand{\Qset}{\ensuremath{\mathcal{Q}}\xspace}
\newcommand{\amap}{\ensuremath{a}\xspace}
\newcommand{\alphamap}[1]{\ensuremath{\alpha_{#1}}\xspace}
\newcommand{\alphaabs}{\ensuremath{\abst\text{-abstraction}}\xspace}
\newcommand{\betamap}[1]{\ensuremath{\beta_{#1}}\xspace}

\definecolor{ck}{RGB}{184, 84, 80}
\definecolor{backbone}{RGB}{108, 142, 191}
\definecolor{system}{RGB}{215, 155, 10}
\definecolor{cochain}{RGB}{150, 115, 166}

%% file: sections/intro.tex
% !TEX root =  ../main.tex
\section{Introduction}\label{sec:introduction}
\emph{\enquote{The Book of Why}} \cite{pearl2018book} elects \emph{causality} as the key to overcoming the limit of purely predictive artificial intelligence 
\blfootnote{Authors' contributions breakdown in back matter. Correspondence to: $\text{<gabriele.dacunto@uniroma1.it>}$ and $\text{<cbattiloro@hsph.harvard.edu>}$.}{(AI)}.
This argument recently found mathematical support in the work of \citet{richens2024robust}, who gave evidence that robustness to distributional shifts of AI agents is conditioned on the learning of an approximate subjective causal model.
The \emph{structural causal model} (SCM) framework \cite{pearl2009causality} is a gold standard for modeling \emph{cause-effect} relationships in complex systems.
Informally, a (probabilistic) SCM is made by causal variables and variable-specific sources of noise, together with structural equations -- to be read as assignments, like in physics -- determining how each variable is causally influenced by others (e.g., the test score of a student depends on hours of study, motivation, and some randomness).
An SCM induces the so-called \enquote{ladder of causations} \cite{pearl2018book}: \faIcon{eye} the \emph{associational} layer related to factual information (\emph{seeing}), \faIcon{hammer} the \emph{interventional} layer related to the effects of actions (\emph{doing}), and \faIcon{brain} the \emph{counterfactual} layer related to imagine the effect of an action, given that something else occurred (\emph{retrospection}). 
The ultimate goal of \emph{causal AI} is to empower AI systems with such a ladder of causation for robust and trustworthy decision-making.
This work pushes in the same direction.\\
\noindent\textbf{Motivation.} We are driven by a philosophically simple yet technically unexplored concept: \textit{any causal model is an imperfect and subjective representation of the world, and it cannot be severed from the network of relations the subject is immersed in}. As such, our work somehow starts where \citet{richens2024robust} end, as they provided a (specific) formalization of the subjectiveness of causal models without, however, formalizing their dependency on the subject's relationships. In philosophy, this concept has been extensively debated: manipulability theories view causes as \enquote{handles} for effecting change, tying them to subjective agency \cite{woodward2001manipulability}; pluralist approaches suggest multiple, context-dependent concepts of causation \cite{psillos2010causalpluralism}; the actor-network theory of Latour and others posits that everything in the social and natural worlds exists in constantly shifting networks of relationships, and nothing exists outside those relationships \cite{latour2007actortheory}. A motivating example for our work is \emph{agentic AI}, a frontier paradigm pushing for autonomous AI agents -- the subjects -- to collaborate in solving complex tasks. 
However, in our work, the notion of \enquote{subject} takes on a broader meaning and could represent, for example, the resolution by which a phenomenon is studied; the sensor that detects pollutants in a specific geographical area, different from that of other sensors; the trading book of a bank seen as a proxy for investment strategy.
In all of the above cases, we advocate that each subject likely -- and, arguably, hopefully -- develops a subjective SCM, these SCMs are interconnected, and their interplay can benefit the subject and the entire network.
Inherently, this paradigm, which we term \emph{relativity of causal knowledge} and whose core object is \emph{relative causal knowledge} (RCK), is tied to the concept of \emph{perspective}: subjects cannot interact by detaching from their world representation. In our framework, asking for a unique \enquote{true} causal description of a system is an ill-posed question, as the very same notion of causality is inherently relative. However, it is important to highlight that the relativity of causal knowledge is different from the \textit{relativism} (in its philosophical meaning) of causal knowledge: \textit{we do not undermine the meaning of things, we question its description as a monolithic object}.\\
\noindent\textbf{Related Works.} RCK has its roots in \emph{category theory} \cite{mac2013categories} and \emph{sheaf theory} \cite{serre1955faisceaux}, and is inspired by Grothendieck's notion of relativism in the context of mathematics \cite{mclarty2003rising}.
Grothendieck revolutionized the understanding of mathematical structures by shifting the focus from individual objects to the relationships between them, as expressed through morphisms. 
This perspective led to a flexible, contextual view of mathematical spaces. 
We draw a parallel to Grothendieck's paradigm shift by treating SCMs as inherently contextual and interconnected rather than isolated entities. 
Just as Grothendieck’s use of sheaf theory facilitates local-to-global transitions in mathematics, we employ network sheaves--first-order cellular sheaves \cite{curry2014sheaves}-- to investigate how causal knowledge, i.e., a set of interventional and observational probability measures, is transferred and transformed across different subjects and their perspectives.
Previous work focused on a functorial characterization of SCM through mappings between the causal structure and the (discrete) distributions of causal variables \cite{jacobs2019causal}.
Conversely, we provide a category-theoretic treatment of SCMs -- and related interventions -- focusing on morphisms between probability spaces, linking the SCM to the category of convex spaces of probability measures \cite{fritz2009convex}. 
Similar definitions to \Cref{def:SCM_meas,def:scm_fun} appear in the concurrent work \cite{d2025causal}.\\
Relations among SCMs have been investigated over time.
The \emph{transportability} problem \cite{pearl2011transportability,bareinboim2016causal} addresses the transfer of causal knowledge from one environment (the source) to another (the target) under assumptions on \emph{(i)} the knowledge of the underlying causal structure, typically represented by causal graphs, and \emph{(ii)} the types and targets of intervention. 
In \emph{causal transfer learning} \cite{zhang2015multi,rojas2018invariant,magliacane2018domain}, several studies tackle the challenge of transferring causal knowledge from source domains to enhance performance in a target domain (i.e., domain adaptation). 
Next, \emph{equivalence} of SCMs \cite{,beckers2021equivalent,verma2022equivalence} aims at identifying equivalent (sub)structures to provide insights into how different systems share common causal relationships.
Particularly relevant to our work is the theory of \emph{causal abstraction} (CA) \cite{rubenstein2017causal,beckers2019abstracting}.
CA formalizes the mappings between SCMs describing the same system at different levels of granularity.
In this paper, we will work under the \alphaabs framework proposed by \citet{rischel2020category}.
The latter is convenient to us since \emph{interventional consistency} (IC) is neatly separated from the definition of the CA.\\
\begin{figure}[t!]
    \centering
    \includegraphics[width=1\linewidth]{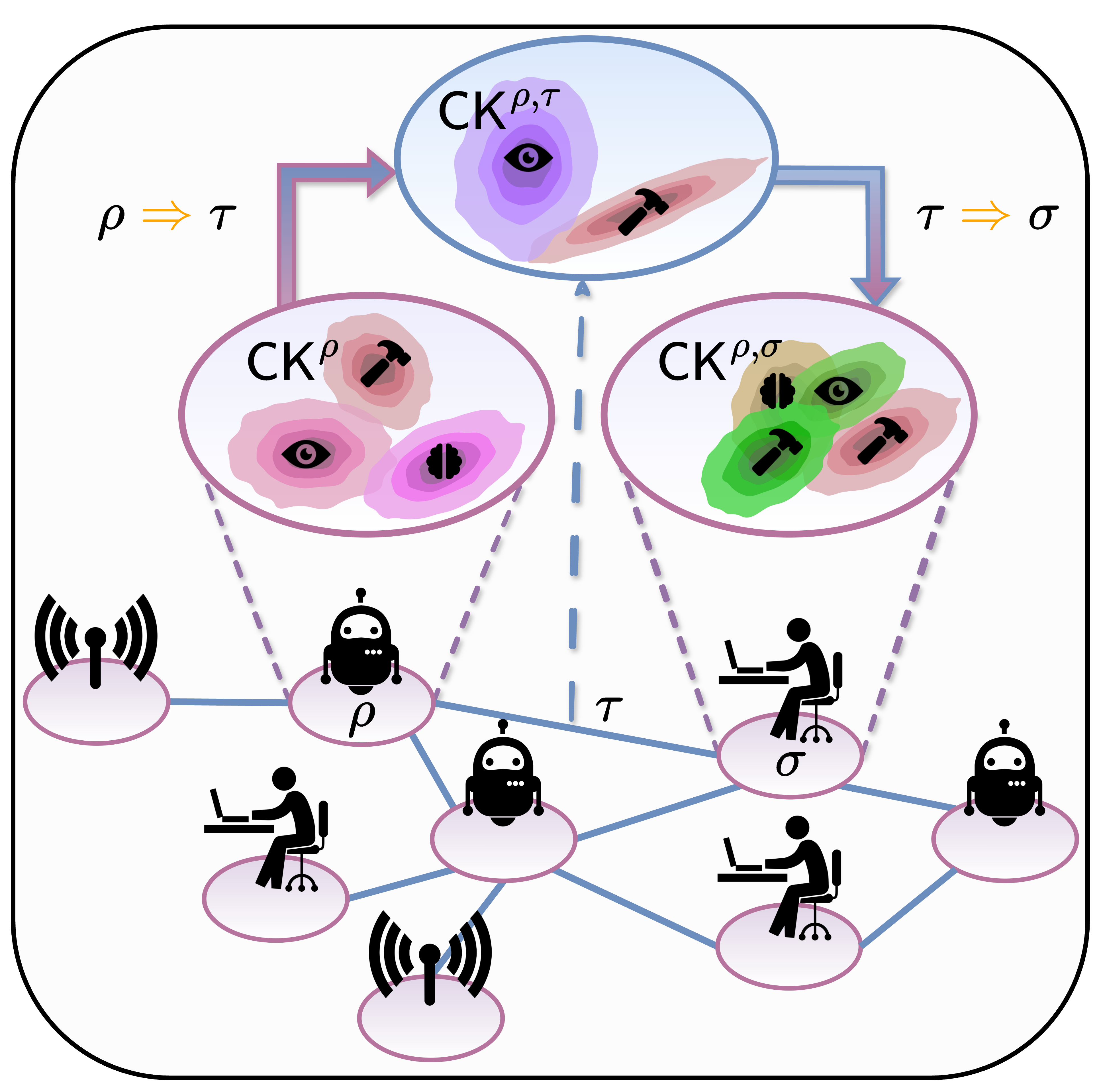}
    \caption{The \emph{relativity of causal knowledge} states that causal knowledge (CK) is subjective and interconnected rather than objective and isolated. 
    Multiple subjects of/in the same system will develop multiple and different instances of CK describing the system. Informally, CK can be seen as a set of probability measures corresponding to \faIcon{eye} seeing, \faIcon{hammer} doing, and imagining \faIcon{brain}.
    The CK $\mathsf{CK}^{\rho}$ of subject $\rho$ is fully accessible exclusively by $\rho$. 
    As such, another subject $\sigma$ can only access the \emph{relative causal knowledge} (RCK) $\mathsf{CK}^{\rho, \sigma}$, i.e., the CK of $\rho$ from the perspective of $\sigma$. 
    There is a link $\tau$ between $\rho$ and $\sigma$, i.e., they can communicate, if their CK admits a common interventionally consistent CA acting as \emph{backbone space}. 
    As such, the RCK $\mathsf{CK}^{\rho, \sigma}$ is obtained by first transporting $\mathsf{CK}^{\rho}$ on $\tau$, obtaining a more abstract $\mathsf{CK}^{\rho, \tau}$, and then  $\mathsf{CK}^{\rho, \tau}$ on $\sigma$. 
    Interestingly, subjects that are not directly connected can still access some RCK if there is a path of links among them, but it would be first \enquote{filtered} by the perspective of all the other subjects on the path. 
    We elegantly implement RCK using a category-theoretic approach resulting in novel mathematical objects: \emph{The network sheaf and cosheaf of causal knowledge.}}
    \label{fig:enter-label}
\end{figure}
\noindent\textbf{Contributions.}
A number of results are established to pose the formal definition of RCK.
\emph{First}, we introduce the category of SCMs, viz. \SCMcat, whose objects are SCMs -- expressed as functors -- and morphisms are natural transformations.
We further characterize hard \cite{pearl2009causality} and soft \cite{eberhardt2007interventions} interventions in the latter category, proving that the set of entailed observational and interventional probability measures is closed under a convex combination operation (cf. \Cref{th:convex_comb_prob_meas}).
\emph{Second}, aiming at IC, we recast the \alphaabs in \SCMcat, proving that IC CA morphism corresponding to endogenous variables is valid in the category of convex spaces of probability measures, viz. \CSprob (cf. \Cref{th:ca_affine_functions}).
Hence, we establish the existence of a functor encoding non-intervened SCMs into objects in \CSprob, representing the SCMs' causal knowledge.
\emph{Third}, leveraging the encoding functor, we define network sheaf and cosheaf of causal knowledge, finally posing the formal definition of RCK.\\
\noindent\textbf{Impact.}
Our proposed network sheaves and cosheaves are particular instances of more general network sheaves and cosheaves in \CSprob. 
Our work ultimately emphasizes their relevance to applications in (but not limited to) AI/ML. 
In particular, RCK and network sheaf and cosheaf of causal knowledge represent new objects to be investigated, adding to the cellular sheaves valued in Abelian categories, such as those on vector spaces, and Hilbert spaces \cite{curry2014sheaves,hansen2019toward}, and the non-Abelian case of sheaves of lattices \cite{ghrist2022cellular}. 
Overall, our work opens three major areas of research: \emph{cohomology theory}, \emph{Hodge(-like) theory}, and \emph{learning theory} for network sheaves and cosheaves of causal knowledge. \\
\noindent\textbf{Notation.}
Sets and collections are uppercase calligraphic, $\mathcal{A}$. 
The set of integers from $1$ to $n$ is $[n]$.
Given a scalar $a$, we denote by $\bar{a}=1-a$.
A set \myexogenousvals equipped with a $\sigma$-algebra $\Upsilon$ gives a measurable space $(\myexogenousvals, \Upsilon)$.
A measurable space $(\myexogenousvals, \Upsilon)$ together with a probability measure $\mu$, i.e., such that $\mu(\myexogenousvals)=1$, gives a probability space $(\myexogenousvals, \Upsilon, \mu)$.
Given two measurable spaces $(\myexogenousvals, \Upsilon)$ and $(\myendogenousvals, \Omega)$, a measurable map $\varphi:\myexogenousvals \rightarrow \myendogenousvals$, and a probability measure $\mu$ over $(\myexogenousvals, \Upsilon)$, we denote by $\varphi(\mu)\coloneqq \mu \,\circ\, \varphi^{-1}$ the pushforward measure over $(\myendogenousvals, \Omega)$.
The domain of a function is $\mathcal{D}[]$. 
The indicator function is $\indicatorf{\mathcal{A}}{A}$, $1$ if $A \in \mathcal{A}$, $0$ otherwise.
\begin{remark}
A less technical but comprehensive description together with practical examples can be found in Appendix \ref{app:disc&ex}. This description is designed to be useful either before or after the reader has parsed the main body of the paper.
\end{remark}

%% file: sections/scm_cs.tex
% !TEX root =  ../main.tex
\section{Categorical SCM and causal knowledge}\label{sec:cat_scm_ck}
Category theory \cite{mac2013categories} is a branch of pure mathematics that studies abstract structures and their relationships through objects and morphisms, focusing on how they compose and interact.
A \emph{category} \Ccat is composed of objects having a certain structure (e.g., measurable spaces, vector spaces) and arrows (morphisms) between them preserving the structure (e.g., measurable maps, linear maps) and satisfying certain axioms (cf. \Cref{def:category}).
Objects and morphisms form the collections $\mathcal{C}_0$ and $\mathcal{C}_1$, respectively.
Given \Ccat, the opposite category $\Ccat^\mathrm{op}$ has the same objects as \Ccat and the same, but reversed, arrows.
Taking the subcollections $\mathcal{S}_0\subseteq \mathcal{C}_0$ and $\mathcal{S}_1\subseteq \mathcal{C}_1$ we can form a \emph{subcategory} of \Ccat.
Given two categories \Ccat and \Dcat, we can have an arrow between them, namely a \emph{functor} $F:\Ccat \rightarrow \Dcat$.
Functors define mappings between categories in a consistent way (cf. \Cref{def:functor}).
Notably, functors cannot destroy relations existing in \Ccat when mapping to \Dcat.
Given two functors $F$ and $G$ from \Ccat to \Dcat, we can have an arrow $\eta: F \rightarrow G$ between them called \emph{natural transformation}, which is a peculiar arrow as it is invariant w.r.t. morphisms between the mapped objects in target category \Dcat (cf. \Cref{def:nat_transf}).  
Functors can be arranged in categories whose objects are functors and morphisms are natural transformations.
More details in \Cref{app:background}.\\
We look at the probabilistic SCM over continuous random variables, denoted by \scm{}, as a subjective and imperfect world representation.
We consider the \emph{Markovian setting} \cite{pearl2009causality}, thus \scm{} entails a \emph{directed acyclic graph} (DAG), viz., $G^{\scm{}}$.
The nodes of $G^{\scm{}}$ correspond to the \emph{endogenous} variables $\myendogenous$, that is, the causal variables on which we can intervene on.
The edges of $G^{\scm{}}$ are determined by structural functions $\myfunctional \coloneqq\{f_1, \ldots, f_n\}$ determining the value of each causal variable $X_i$, $i\in[n]$, in terms of a set of \emph{parents}, viz. $\parents_i \subseteq \myendogenous \setminus \{X_i\}$, and a node-specific \emph{exogenous} variable, $Z_i \in \myexogenous$.
Denote by $\myexogenous^{\mathcal{A}_i} \subseteq \myexogenous \setminus \{ Z_i\}$ the set of exogenous variables corresponding to the ancestors of $X_i$, where $\ancestors_i \subseteq [n] \setminus \{i\}$.
\myfunctional induces a set of mixing functions $\mymixing\coloneqq\{m_1, \ldots, m_n\}$, such that the values of the endogenous random variables are equivalently expressed as $x_i=m_i\left(\{z_j\}_{j\in \ancestors_i}, z_i\right)$, $\forall \; i \in [n]$.\\
Accordingly, we can define \scm{} as a triple $\langle (\myexogenousvals,\, \Upsilon, \zeta), \, (\myendogenousvals,\, \Omega, \chi)\, , \mymixing \rangle$ made of the probability space of the exogenous $(\myexogenousvals,\, \Upsilon, \zeta)$, the probability space of the endogenous $(\myendogenousvals,\, \Omega, \chi)$, and measurable mixing functions \mymixing.
\begin{definition}[Measure-theoretic SCM]\label{def:SCM_meas}
    A Markovian SCM \scm{} is a triple $\langle (\myexogenousvals,\, \Upsilon, \zeta), \, (\myendogenousvals,\, \Omega, \chi)\, , \mymixing \rangle$  where:
    \begin{squishlist}
        \item $(\myexogenousvals,\, \Upsilon, \zeta)$ is a probability space associated with exogenous variables \myexogenous. Specifically, it consists of the product probability measure $\zeta=\zeta_1 \times \ldots \times \zeta_n$ on the product measurable space $(\myexogenousvals,\, \Upsilon)=(\myexogenousvals_1 \times \ldots \times \myexogenousvals_n,\, \Upsilon_1 \otimes \ldots \otimes \Upsilon_n)$.
        The probability measure is such that, for each $\mathcal{W}_1 \in \Upsilon_1, \ldots, \, \mathcal{W}_n \in \Upsilon_n$, we have 
        \begin{equation}
            \zeta_1 \times \ldots \times \zeta_n (\mathcal{W}_1 \times \ldots \times \mathcal{W}_n)=\zeta_1(\mathcal{W}_1) \times \ldots \times \zeta_n(\mathcal{W}_n)\,;
        \end{equation}
        \item $(\myendogenousvals,\, \Omega, \chi)$ is a probability space associated with endogenous variables \myendogenous consisting of a joint probability measure $\chi$ on the product measurable space $(\myendogenousvals,\, \Omega)=(\myendogenousvals_1 \times \ldots \times \myendogenousvals_n,\, \Omega_1 \otimes \ldots \otimes \Omega_n)$;
        \item $\mymixing\coloneqq\{m_1, \ldots, m_n\}$ is a collection of measurable mixing functions allowing us to recursively rewrite the causal assignments only in terms of the exogenous variables. 
        Accordingly, the joint probability measure $\chi$ factorizes as 
        \begin{equation}\label{eq:scm_obs_push}
            \chi = \bigtimes_{i=1}^n m_i\left(\mu_i \left( \myexogenousvals_i \times \myexogenousvals^{\ancestors_i} \right) \right)\,;
        \end{equation}
        where (i) $\myexogenousvals^{\ancestors_i}=\bigtimes_{j \in \ancestors_i} \myexogenousvals_j$; 
        (ii) $\mu_i$ is a probability measure on the product measurable space $\left( \myexogenousvals_i \times \myexogenousvals^{\ancestors_i},\, \Upsilon_i \otimes \Upsilon^{\ancestors_i} \right)$, with $\Upsilon^{\ancestors_i}=\bigotimes_{j \in \ancestors_i} \Upsilon_j$.
    \end{squishlist}
\end{definition}

\medskip
\Cref{app:examples} provides an example for the case of linear SCM with additive noise.
An important property of \mymixing that holds for the main classes of Markovian SCMs, such as additive noise and post-nonlinear models \cite[Chapter~4]{peters2017elements}, is invertibility (cf. \Cref{app:invertibility}).
We will leverage such a property to prove that the proposed \SCMcat preserves expressiveness.
At this point, we can leverage \Cref{def:SCM_meas} to give a functorial definition of \scm{}.
Consider the following categories, \I and \Prob, respectively.
The former is a small category made only of two objects, a source node and a target node, and a unique arrow from the source to the target node.
Specifically, \I has shape $\bullet \rightarrow \bullet$.
The latter category instead has as objects probability spaces and as morphisms measurable maps.
The functorial representation follows by viewing \scm{} as an arrow between \I and \Prob, assigning \emph{(i)} to the source node in \I the probability space associated with exogenous variables; \emph{(ii)} to the target node the probability space associated with endogenous variables; 
and \emph{(iii)} to the only arrow in \I the collection of measurable maps \mymixing.
\begin{definition}[Category-theoretic SCM]\label{def:scm_fun}
    An SCM is a functor $\scm{}: \I \rightarrow \Prob$ defined as follows
    \begin{equation}
        \centering
        \begin{tikzpicture}[]
    
        \node (I) at (0, 2.25) {\I};
        \node (P) at (3, 2.25) {\Prob};
        
        \node[circle, draw, fill,inner sep=1pt] (A) at (0, 1.5) {};
        \node[circle, draw, fill,inner sep=1pt] (B) at (0, 0) {};
        \node (A1) at (-0.3, 1.5) {I};
        \node (B1) at (-0.3, 0) {$I^\prime$};
        \node (C) at (3, 1.5) {$(\myexogenousvals,\, \Upsilon, \zeta)$};
        \node (D) at (3, 0) {$(\myendogenousvals,\, \Omega, \chi)$};
    
        \coordinate (A1shift) at ([yshift=-5pt]A);
        \draw[->,shorten >=2pt] (A1shift) -- (B);
    
        \coordinate (Ishift) at ([xshift=10pt]I);
        \coordinate (Pshift) at ([xshift=-20pt]P);
        \draw[->] (Ishift) -- node[above] {\scm{}} (Pshift);
        \draw[->] (C) -- node[right] {\mymixing} (D);
        \end{tikzpicture}
        \label{fig:scmfunctor}
    \end{equation}
\end{definition}
Please, refer to \Cref{app:examples} for an example in the case of linear SCM with additive noise.
At this point, we can define a category of SCMs whose objects are functors as in \Cref{def:scm_fun}, and whose morphisms are natural transformation between the functors.
\begin{definition}\label{def:SCMcat}
    The category of SCMs, namely \SCMcat, consists of (i) functors $\scm{}: \I \rightarrow \Prob$ as objects; and (ii) natural transformations $\eta: \scm{} \rightarrow \scm{\prime}$ as morphisms, such that:
    \begin{squishlist}
        \item for each $I$ in \I, a measurable map $\eta_I: \scm{}(I) \rightarrow \scm{\prime}(I)$ in \Prob, called component at $I$;
        \item for the unique morphism $f:I \rightarrow I^\prime$ in \I, the following commutes:
            \begin{equation}\label{eq:nat_transf_SCMcat}
                \begin{tikzcd}[row sep=1.5cm, column sep=1.5cm]
                    \scm{}(I) \arrow[r, "\scm{f} = \mymixing"] \arrow[d, "\eta_I"'] & \scm{}(I^\prime) \arrow[d, "\eta_{I^\prime}"] \\
                    \scm{\prime}(I) \arrow[r, "\scm{\prime^f} = \mymixing^\prime"'] & \scm{\prime}({I^\prime)}
                \end{tikzcd}
                \begin{tikzpicture}[overlay]
                    \draw[dashed, rounded corners] (-4,-1.5) rectangle (-2.8,1.5);
                    \draw[dashed, rounded corners] (-1.2,-1.5) rectangle (0,1.5);
                    \node at (-3.4,-1.8){Exogenous };
                    \node at (-0.6,-1.8){Endogenous };
                \end{tikzpicture}
            \end{equation}  
        \end{squishlist}
\end{definition}
In \Cref{eq:nat_transf_SCMcat}, we added dashed rectangles to remark that the functor images \emph{(i)} on the left correspond to the probability spaces of the exogenous variables of \scm{} and $\scm{\prime}$;
\emph{(ii)} on the right, to probability spaces of the endogenous variables.\\
The knowledge of an SCM allows us to run \emph{interventions} to act or to simulate \enquote{new worlds}, that is, obtain new post-interventional distributions.
This enables the second and third layers of the ladder of causation \cite{pearl2018book} described in \Cref{sec:introduction}. 
In our work, we consider \emph{hard} and \emph{soft} interventions.\\
A hard intervention $\operatorname{do}\left(\{X_i = x_i\}_{X_i \in \widetilde{\mathcal{X}}}\right)$, where $\widetilde{\mathcal{X}} \subseteq \myendogenous$, replaces each assignment function $f_i$ corresponding to $X_i \in \widetilde{\mathcal{X}}$ with the constant $x_i$, thus generating a new post-interventional SCM, viz. $\scm{}_\hard$.
We associate a hard intervention with a collection $\mathcal{F}_\hard$ such that $f_i^\hard=x_i$, for all $X_i \in \widetilde{\mathcal{X}}$; and $f_j^\hard=f_j$, for all $X_j \in \myendogenous \setminus \widetilde{\mathcal{X}}$.
Graphically, a hard intervention mutilates $G^{\scm{}}$ by removing the incoming edges of the variables in $\widetilde{\mathcal{X}}$.
Consequently, according to the truncated product formula \cite{pearl2009causality}, an intervention entails a post-interventional distribution $P({\myendogenous}\mid \hard)= \prod_{j \in \myendogenous \setminus \widetilde{\mathcal{X}}} P(X_j \mid \parents_i, Z_i)$ (evaluated at $\{X_i = x_i\}$).\\
In \SCMcat, the post-interventional SCM is the functor mapping \emph{(i)} the source node in \I to $(\myexogenousvals,\, \Upsilon, \zeta)$, as done by \scm{}; \emph{(ii)} the target node to $(\myendogenousvals,\, \Omega, \chi)_\mathcal{H}$, where $\chi_j$ degenerates to a point measure $\indicatorf{x_j}{X_j}$; \emph{(iii)} and the unique arrow in \I to the measurable maps $\mymixing_\hard \coloneqq \{m_1^\hard, \ldots, m_n^\hard\}$.
The components of $\mymixing_\hard$ are \emph{(i)} constant functions $m_i^\hard=x_i$ for $X_i \in \mathcal{\myendogenous}$,
and \emph{(ii)} $m_j^\hard = m_j^\hard(Z_j \cup \myexogenous^{\widetilde{\ancestors}_j})$ where $\myexogenous^{\widetilde{\ancestors}_j}$ is the set of exogenous corresponding to the ancestors of $X_j$ that are not screened by the intervention (cf. \Cref{app:examples}).
At this point, in the same spirit of the truncation formula above, the post-interventional probability measure reads as 
\begin{equation}\label{eq:scm_hint_push}
    \chi_{\mathcal{H}} = \bigtimes_{X_j\in \myendogenous\setminus\widetilde{\mathcal{X}}} m_j^\hard\left(\mu_j \left( \myexogenousvals_j \times \myexogenousvals^{\widetilde{\ancestors}_j} \right) \right)\,.
\end{equation}
Running hard interventions on SCMs is not always possible, and in certain cases, this way of intervening is unethical \cite{eberhardt2007interventions}.
Therefore, a more general notion of intervention has been considered over the past. 
Indeed, the mutilation of the DAG is not the only possible informative intervention.
Rather, we can be interested in simply modifying the causal mechanisms, without removing any incoming causal relations.
Such a family of interventions is called soft.\\
The soft intervention generates a post-interventional $\scm{}_\soft$ by substituting \myfunctional with $\myfunctional_\soft$, where each $f_i$ associated with the intervened variables $X_i \in \widetilde{X}$ is replaced by another function $\widetilde{f}_i$, and the rest is unchanged.
Hence, the soft intervention can change the functional form of an SCM.
In principle, soft interventions can potentially add new causal relations.
However, in our work, we consider only soft interventions that do not alter the parent set of the endogenous variables.
At this point, $\myfunctional_\soft$ induces a new collection of mixing functions $\mymixing_\soft \coloneqq \{m^\soft_1, \ldots, m^\soft_n\}$ (cf. \Cref{app:examples}), such that the post-interventional probability measure reads as
\begin{equation}\label{eq:scm_sint_push}
    \chi_{\soft} = \bigtimes_{X_i \in \myendogenous} m_i^\soft\left(\mu_i \left( \myexogenousvals_i \times \myexogenousvals^{\ancestors_i} \right) \right) \,.
\end{equation}
The resulting $\scm{}_\soft$ is an object in \SCMcat mapping \emph{(i)} the source node in \I to $(\myexogenousvals,\, \Upsilon, \zeta)$, as done by \scm{}; \emph{(ii)} the target node to $(\myendogenousvals,\, \Omega, \chi)_\mathcal{\soft}$, where $\chi_\soft$ follows \Cref{eq:scm_sint_push}; and \emph{(iii)} the unique arrow in \I to the measurable map $\mymixing_\soft$.\\
An important point we must ensure is that the proposed category-theoretic SCM formulation preserves the expressiveness: given either a hard or soft intervention \intervention, we have a suitable morphism in \SCMcat from \scm{} to $\scm{}_\intervention$.
This is ensured by the following.
\begin{restatable}{lemma}{interventionlemma}\label{lem:intervention_prob}[Interventions in \SCMcat]
    Given (i) \scm{} as in \Cref{def:scm_fun}; and (ii) the collection of measurable maps $\mathcal{F}_\intervention\coloneqq \{f^i_1, \ldots, f^i_n\}$;
    an intervention in \SCMcat is a natural transformation $\eta^\intervention\coloneqq\langle \catidentity_{\scm{}(I)},\, \intervention \rangle$, with $\intervention = \mymixing_\intervention \circ \mymixing^{-1}$ such that the following holds 
\end{restatable}
\begin{equation}\label{eq:nat_transf_inter_SCMcat}
    \begin{tikzcd}[row sep=1.5cm, column sep=1.5cm]
        \scm{}(I) \arrow[r, "\mymixing"] \arrow[d, "\catidentity_{\scm{}(I)}"'] & \scm{}(I^\prime) \arrow[d, "\intervention"] \\
        \scm{}_{\intervention}(I) \arrow[r, "\mymixing_\intervention"'] & \scm{}_\mathcal{H}({I^\prime)}
    \end{tikzcd} \Longrightarrow  \mymixing_\intervention = \intervention \circ \mymixing\,. 
\end{equation}
\begin{proof}
    See \cref{app:proofs}.
\end{proof}
The proof leverages the fact that the hard intervention acts only on the endogenous, thus the component $\eta^\intervention_I$ mapping $\scm{}(I)$ to $\scm{}_\intervention(I)$ is simply the identity $\catidentity_{\scm{}(I)}$.
Then, for the second component $\eta^\intervention_{I^\prime}$, we leverage invertibility of \mymixing mentioned above.
The commutation in \Cref{eq:nat_transf_inter_SCMcat} follows by construction.
\Cref{eq:nat_transf_inter_SCMcat} highlights that our proposed \SCMcat, is as rich as the canonical SCM framework.
Specifically, starting from a non-intervened SCM \scm{}, we can obtain all possible observational and interventional states of the causal model \scm{} through the application of all the possible hard and soft interventions since \Cref{lem:intervention_prob} guarantees the existence of corresponding measurable maps \intervention.
\begin{remark}
    By leveraging the invertibility of \mymixing, \Cref{lem:intervention_prob} preserves expressiveness not only at the level of probability measures, but also with respect to the values of exogenous and endogenous variables.
    This is a stronger result, as interventional consistency concerns only the distributional level.
    In fact, \SCMcat retains the expressiveness of canonical SCMs also in terms of counterfactual consistency.
    If counterfactual consistency is relaxed, a more general result on the existence of the morphism \intervention can be established via Brenier's polar factorization theorem \cite[Chapter~3]{villani2021topics}, without requiring the invertibility of \mymixing, as shown in \cite{d2025causal}.
\end{remark}
\begin{definition}[Causal knowledge]\label{def:causal_knowledge}
    The causal knowledge \CK{\scm{}} entailed by \scm{} is the subcategory of \SCMcat whose objects are \scm{} together with its intervened states $\{\scm{}_\intervention\}$ generated by the application of all the possible interventions $\{\intervention\}$, and whose morphisms are the natural transformations $\eta^\intervention$ between these objects in \SCMcat.  
\end{definition}
Consequently, \CK{\scm{}} corresponds to \emph{(i)} a product distribution over the exogenous variables together with the identity morphism, representing the component $\eta_I^\intervention$ of the natural transformations; and \emph{(ii)} the set of probability measures over the endogenous as given in \Cref{eq:scm_obs_push,eq:scm_hint_push,eq:scm_sint_push} together with the morphisms \hard and \soft, representing the natural transformation component $\eta_{I'}^\intervention$.
Consider now two different non-intervened SCMs, namely \scm{} and $\scm{\prime}$, and their corresponding \CK{\scm{}} and $\CK{\scm{\prime}}$, respectively.
Additionally, suppose that \scm{} and $\scm{\prime}$ are not isolated entities; rather, they are immersed in a certain network where other non-intervened SCMs exist.
We are interested in relating \CK{\scm{}} and $\CK{\scm{\prime}}$ within the network in a way that is both \emph{causally and category-theoretically} consistent.
From now on, we will focus on the endogenous layer, since exogenous variables are latent.
Hence, when we say \CK{\scm{}}, with slight abuse of notation we refer to the observational and interventional probability measures over the endogenous.

%% file: sections/causal_knowledge.tex
% !TEX root =  ../main.tex
\section{Encoding causal knowledge with convex spaces}\label{sec:encoding_ck}
Let us consider the measurable space $(\myendogenousvals,\, \Omega)$ associated with the endogenous variables, where $\Omega$ is a $\sigma$-algebra over \myendogenousvals.
The set of probability measures on $(\myendogenousvals,\, \Omega)$, namely $\Delta_{(\myendogenousvals,\, \Omega)}$ is a convex space, subset of the vector space $\reall^\Omega$.
Following \cite{fritz2009convex}, we define the latter convex space as follows.
\begin{restatable}{lemma}{convexspaceprob}\label{lem:convexspace_prob}[Convex space of probability measures, $\langle \Delta_{(\myendogenousvals,\, \Omega)}, cc_{\lambda} \rangle$]
    The convex space of probability measures, namely $\langle \Delta_{(\myendogenousvals,\, \Omega)}, cc_{\lambda} \rangle$, is given by the set $\Delta_{(\myendogenousvals,\, \Omega)}$ of probability measures $\chi$ on $(\myendogenousvals,\, \Omega)$ together with a convex combination operation defined by 
    \begin{equation}\label{eq:ccl}
        cc_{\lambda}(\chi_1, \chi_2)(\mathcal{O}) \coloneqq \lambda \chi_1(\mathcal{O}) + \bar{\lambda} \chi_2(\mathcal{O})\,, 
    \end{equation}
    for all $\mathcal{O} \in \Omega$, with $\lambda \in [0,1]$.
\end{restatable}
\begin{proof}
    See \cref{app:proofs}.
\end{proof}
Convex spaces $\langle \Delta_{(\myendogenousvals,\, \Omega)}, cc_{\lambda} \rangle$ are the objects of the category of convex spaces of probability measures, namely \CSprob, that we introduce below.
\begin{restatable}{lemma}{csprobcat}    
    The category \CSprob has convex spaces $\langle \Delta_{(\myendogenousvals,\, \Omega)}, cc_{\lambda} \rangle$ as objects, and affine measurable maps -- i.e., measurable maps commuting with $cc_\lambda$ -- as morphisms.
\end{restatable}
\begin{proof}
    See \cref{app:proofs}.
\end{proof}
It turns out that \CK{\scm{}} is closed w.r.t. the convex combination operation in \Cref{lem:convexspace_prob}.
\begin{restatable}{theorem}{convexcomb}\label{th:convex_comb_prob_meas}
    Every convex combination of probability measures corresponding to a causal knowledge \CK{\scm{}} is a valid soft-interventional probability measure for \CK{\scm{}}.
\end{restatable}
\begin{proof}
    See \cref{app:proofs}.
\end{proof}
Hence, in light of \Cref{th:convex_comb_prob_meas}, it is legitimate to question whether we can establish a functorial encoding of \scm{} in a convex space.
If on objects, the encoding seems straightforward, on morphisms, we have to be careful.
Indeed, given \scm{} and $\scm{\prime}$, the natural transformation $\eta$ in \Cref{def:SCMcat} tells us nothing about the existence of natural transformations between the intervened states of \scm{} and $\scm{\prime}$.
If the latter natural transformations do not exist in \SCMcat, an encoding functor will fail in mapping $\eta$ to an affine morphism between the convex spaces corresponding to \scm{} and $\scm{\prime}$, since such a morphism does not exist in \CSprob.
Conversely, if the intervened states of \scm{} and \scm{\prime} are related by a natural transformation in \SCMcat, meaning that they are IC, the affine morphism will exist in \CSprob.\\
The latter observation naturally links our work to the theory of \emph{causal abstraction} (CA), electing CA as the necessary formalism for relating causal knowledge.
Specifically, in the following, we build upon the \alphaabs introduced by \cite{rischel2020category}.
Given a micro-level \scm{} and a macro-level $\scm{\prime}$, an \alphaabs is a triple $\boldsymbol{\alpha}\coloneqq \langle \Rset, \amap, \alphamap{} \rangle$ where \emph{(i)} \Rset is a set of endogenous variables in \scm{} that are abstracted to the macro-level \emph{(ii)} structurally via the surjective map $\amap:\Rset\rightarrow \myendogenous^\prime$, and \emph{(iii)} functionally by  $\alphamap{}: \mathcal{D}[\Rset] \rightarrow \mathcal{D}[\myendogenous^\prime]$.
Here, for each $X_i^\prime \in \myendogenous^\prime$, we have a surjective function mapping the values of the micro-level to the macro-level, viz.  $\alphamap{X_i^\prime}: \mathcal{D}[\amap^{-1}\left(X_i^\prime\right)] \rightarrow \mathcal{D}[X_i^\prime]$. 
Given an \alphaabs between \scm{} and $\scm{\prime}$, we say that \abst is IC if, \emph{for any intervention} \intervention on the endogenous $\myendogenous^\prime_\intervention \subseteq \myendogenous^\prime$ and for any set of targets $\mathcal{T}^\prime \subseteq \myendogenous^\prime \setminus \myendogenous^\prime_\intervention$, we can obtain the values in $\mathcal{D}[\mathcal{T}^\prime]$ starting from those in $\mathcal{D}[\amap^{-1}\!\left(\myendogenous^\prime_\intervention \right)]$ in two alternative ways: 
\emph{(i)} by computing the values in $\mathcal{D}[\amap^{-1}\!\left(\mathcal{T}^\prime \right)]$ at the micro-level, then abstracting via \alphamap{\mathcal{T}^\prime}; or \emph{(ii)} by abstracting via \alphamap{\myendogenous^\prime_\intervention} to $\mathcal{D}[\myendogenous^\prime_\intervention]$ and then computing the values in $\mathcal{D}[\mathcal{T}^\prime]$ at the macro-level.\\
The \alphaabs manifests in \SCMcat as a natural transformation between the micro-level \scm{} and the macro-level $\scm{\prime}$.
As highlighted in \Cref{def:scm_fun,def:SCMcat}, our category-theoretic formalism highlights the role of the exogenous.
Accordingly, we need to consider the exogenous to properly extend the \alphaabs to \SCMcat.
Specifically, \amap and \alphamap{} have two components, the first for the exogenous, and the second for the endogenous, as formalized below.
\begin{definition}[$\abst$-abstraction in \SCMcat]\label{def:alpha_abstraction_scmcat}
    Given micro- and macro-level SCMs, \scm{} and $\scm{\prime}$, respectively, an \alphaabs is a tuple $\abst = \langle \Rset, \Qset, \amap, \alphamap{} \rangle$, where: (i) $\Rset \subseteq \myendogenous$ is a set of relevant endogenous variables; (ii) $\Qset \subseteq \myexogenous$ is a set of relevant exogenous variables given by the union of the set of exogenous corresponding to the endogenous in \Rset and those corresponding to their ancestors; (iii) $\amap=\langle \amap_\myexogenous, \amap_\myendogenous \rangle$ is a pair of surjective functions mapping sets, $\amap_\myexogenous: \Qset \rightarrow \myexogenous^\prime$ and $\amap_\myendogenous: \Rset \rightarrow \myendogenous^\prime$, respectively; (iv) $\alphamap{}=\langle \alphamap{\myexogenous}, \alphamap{\myendogenous} \rangle$ is a natural transformation composed of measurable functions mapping probability spaces, $\alphamap{\myexogenous}$ for the exogenous and $\alphamap{\myendogenous}$ for the endogenous, respectively.  
\end{definition}
In addition, interventional consistency translates into commutation of diagrams in \SCMcat.
Here, we look at an intervention on the endogenous $\myendogenous_\intervention^\prime \in \myendogenous^\prime$ on the macro-level model \scm{\prime} as a collection of measurable maps, as given in \Cref{lem:intervention_prob}, and denote it by $\intervention^\prime$.
Similarly for the corresponding intervention on the micro-level variables $\amap_\myendogenous^{-1}(\myendogenous_\intervention^\prime) \in \myendogenous$, denoted by $\intervention$.
To aid visualization, we use \emph{(i)} violet for micro-level \scm{} layer, \emph{(ii)} gray for macro-level $\scm{\prime}$ layer, and \emph{(iii)} orange for the \alphaabs.
\begin{definition}[IC $\abst$-abstraction in \SCMcat]\label{def:IC_alpha_abstraction_scmcat}
    An $\abst$-abstraction is IC in \SCMcat if, for all the interventions $\eta^{\intervention^\prime}=\langle \catidentity_{\scm{\prime}(I)}, \intervention^\prime \rangle$ on the macro-level model $\scm{\prime}$, the faces of the following diagram commute
    \tdplotsetmaincoords{0}{0}
    \begin{tikzpicture}[tdplot_main_coords, node distance=2cm]
    
      \node (A) at (0,0,0) {$\scm{}(I)$};
      \node (B) at (3,0,0) {$\scm{}(I')$};
      \node (C) at (0,-2,0) {$\scm{\prime}(I)$};
      \node (D) at (3,-2,0) {$\scm{\prime}(I')$};
      
      \draw[->,Mulberry] (A) -- node[above] {$\mymixing$} (B);
      \draw[->,Melon] (A) -- node[left]  {$\alphamap{\myexogenous}$} (C);
      \draw[->,Melon] (B) -- node[right] {$\alphamap{\myendogenous}$} (D);
      \draw[->,Periwinkle] (C) -- node[below] {$\mymixing^\prime$} (D);
      
      \coordinate (shift) at (4,2,1);
      
      \node (A2) at ($(A)+(shift)$) {$\scm{}_\intervention(I)$};
      \node (B2) at ($(B)+(shift)$) {$\scm{}_\intervention(I')$};
      \node (C2) at ($(C)+(shift)$) {$\scm{\prime}_{\intervention^\prime}(I)$};
      \node (D2) at ($(D)+(shift)$) {$\scm{\prime}_{\intervention^\prime}(I')$};
      
      \draw[->,Mulberry] (A2) -- node[above] {$\mymixing_\intervention$} (B2);
      \draw[->,Melon] (A2) -- node[left]  {$\alphamap{\myexogenous}$} (C2);
      \draw[->,Melon] (B2) -- node[right] {$\alphamap{\myendogenous}$} (D2);
      \draw[->,Periwinkle] (C2) -- node[below] {$\mymixing^\prime_{\intervention^\prime}$} (D2);
      
      \draw[->,dashed,Mulberry] (A) -- node[left, xshift=-5mm] {$\catidentity_{\scm{}(I)}$} (A2);
      \draw[->,dashed,Mulberry] (B) -- node[right, xshift=5mm] {$\intervention$}(B2);
      \draw[->,dashed,Periwinkle] (C) -- node[left, xshift=-5mm] {$\catidentity_{\scm{\prime}(I)}$} (C2);
      \draw[->,dashed,Periwinkle] (D) -- node[right, xshift=5mm] {$\intervention^\prime$} (D2);
    \end{tikzpicture}
\end{definition}
Starting from \Cref{def:IC_alpha_abstraction_scmcat}, exploiting \Cref{th:convex_comb_prob_meas}, we obtain the following.
\begin{restatable}{theorem}{caaffinefunctions}\label{th:ca_affine_functions}
    The component $\alphamap{\myendogenous}$ of \alphamap{} within an IC \alphaabs commutes with $cc_{\lambda}$, thus is affine.
\end{restatable}
\begin{proof}
    See \cref{app:proofs}.
\end{proof}
At this point, denoted by \NI the subcategory of \SCMcat whose objects are non-intervened SCMs, and morphisms are IC \alphaabs.
Exploiting \Cref{th:convex_comb_prob_meas} and \Cref{th:ca_affine_functions}, we have the following.
\begin{restatable}{theorem}{encodingfunctor}\label{th:encoding_functor}
    There exists an IC encoding functor $E: \NI \rightarrow \CSprob$ mapping (i) each non-intervened $\scm{} \coloneqq \langle(\myexogenousvals,\, \Upsilon, \zeta), \, (\myendogenousvals,\, \Omega, \chi)\, , \mymixing \rangle$ to the convex spaces of probability measures $\Delta_{(\myendogenousvals,\, \Omega)}$, and (ii) an IC \alphaabs between \scm{} and \scm{\prime} with its endogenous component \alphamap{\myendogenous}.
\end{restatable}
\begin{proof}
    See \cref{app:proofs}.
\end{proof}

%% file: sections/cellular_sheaf.tex
% !TEX root =  ../main.tex
\section{The network sheaf and cosheaf of causal knowledge}\label{sec:CS_CSprob}
The last step toward our proposed \emph{relativity of causal knowledge} is representing causal knowledge over the endogenous within a network.
To this aim, each node within the network is attached to the causal knowledge entailed by a certain SCM via \Cref{th:encoding_functor}.
Looking at a single node $\rho$, we interpret it as the perspective of the causal knowledge on itself.
Then, when two nodes $\rho$ and $\sigma$ are connected by an edge, say $\tau: \rho \sim \sigma$, they are related by a shared IC \alphaabs living on the edge $\tau$.
The causal knowledge at $\rho$ can be transported via $\tau$ to $\sigma$ giving life to the relative causal knowledge (RCK): \emph{the causal knowledge of $\rho$ from the perspective of $\sigma$.}\\
To make this point clearer, consider two AI agents with their subjective causal models \scm{\rho} and \scm{\sigma} about a certain phenomenon.
By \Cref{th:encoding_functor}, we can map \scm{\rho} and \scm{\sigma} to the corresponding causal knowledge, viz. $\mathsf{CK}^\rho\equiv\CK{\scm{\rho}}$ and $\mathsf{CK}^\sigma\equiv\CK{\scm{\sigma}}$.
The causal model \scm{\rho} admits an IC causal abstraction, viz. \scm{\tau}.
The same abstraction holds for \scm{\sigma}.
Acting on \scm{\tau} as well, the encoding functor provides us with a convex space $\mathsf{CK}^\tau$ which can be interpreted as a \emph{backbone space} on which the $\mathsf{CK}^\rho$ and $\mathsf{CK}^\sigma$ can relate each other.
In particular, the AI agents use the backbone space to \emph{embed} the causal knowledge of the other into their own.
Intuitively, consider that the AI agent $\rho$ \emph{projects} via an abstraction morphism an observational/interventional probability measure onto the backbone space $\mathsf{CK}^\tau$, matching an abstracted measure $\chi^{\rho,\tau}$.
The latter can be observed by the AI agent $\sigma$, who embed $\chi^{\rho,\tau}$ in its own causal knowledge obtaining the relative $\chi^{\rho,\sigma}$, namely the causal probability measure entailed by $\scm{\rho}$ from the perspective of $\scm{\sigma}$.\\
Using the categorical foundations laid in \Cref{sec:cat_scm_ck,sec:encoding_ck}, we formalize RCK through the \emph{network sheaf and cosheaf of causal knowledge}, the latter being the dual construction of the former.
Here, the \enquote{dual} nomenclature is used with a slight abuse of notation to mimic the jargon used in Abelian categories \cite{curry2014sheaves,hansen2019toward}.
These mathematical objects consist of \emph{(i)} a network, \emph{(ii)} convex spaces encoding causal knowledge on nodes and edges, and \emph{(iii)} mappings to move from nodes to edges in the case of the sheaf and vice-versa for the cosheaf. \\
First, we introduce the network as a topological object. 
Recall that the network is shaped by the existence of an IC CA between SCMs.
Such a network is shared by the network sheaf and cosheaf.
\begin{definition}[Network]\label{def:net}
    A (finite) network $G\coloneqq(\mathcal{N}, \mathcal{E})$ consists of: 
    (i) nodes $\rho \in \mathcal{N}$ homeomorphic to a point (0-dimensional open ball), 
    and (ii) edges $\tau \in \mathcal{E}$ homeomorphic to an open interval (1-dimensional open ball).
    The closure of each edge is the edge itself plus the two nodes at its boundaries, whereas the node has its closure as there are no lower-dimensional constituents of the network.
    The face incidence relation induces a partial order \faceincidenceposet on the set of nodes and edges, that is, $\rho \trianglelefteq \tau$ if and only if $\rho$ (node) belongs to the closure of $\tau$ (edge).
\end{definition}
Second, the convex spaces of probability measures are objects in \CSprob.
For the network sheaf, they are dubbed \emph{stalks} and are given by the causal knowledge on the network constituents encoded by the functor $E$.
Consider $\tau: \rho \sim \sigma$.
The stalk at $\rho$ is the image $E(\scm{\rho})$ in \CSprob of the SCM $\scm{\rho}$; similarly for $\sigma$.
The stalk at $\tau$ instead is the image $E(\scm{\tau})$ in \CSprob of the causal abstraction shared by $\rho$ and $\sigma$, viz. $\scm{\tau}$.
The role of the CA is key to relating the causal knowledge at $\rho$ and $\sigma$ as it provides a \emph{backbone space} where the causal knowledge can be compared.
For the network cosheaf instead, the convex spaces are dubbed \emph{costalks}.
Consider again $\tau: \rho \sim \sigma$.
The costalk at $\tau$, viz. $\widehat{E}(\scm{\tau})$, coincides with $E(\scm{\tau})$.
The costalk at $\rho$ instead is a convex space of probability measures $\widehat{E}(\scm{\rho})$, embedded in $E(\scm{\rho})$ as specified in the sequel.\\
Third, the mappings manifest as morphisms in the target category, viz. \CSprob.
Specifically, in the case of the network sheaf, we \emph{project} $E(\scm{\rho})$ onto $E(\scm{\tau})$ via the IC CA map, hereinafter \emph{restriction map} and denoted by $\alphamap{\myendogenous}^{\rho\trianglelefteq\tau}$.
Conversely, in the case of the network cosheaf, we \emph{embed} $E(\scm{\tau})$ into $E(\scm{\rho})$ obtaining $\widehat{E}(\scm{\rho})$ via an affine measurable map $\betamap{\myendogenous}^{\rho\trianglelefteq\tau}$, hereinafter \emph{extension map}.
At this point, we can formally define the \emph{embedded costalk of causal knowledge} as the convex space $\widehat{E}(\scm{\rho})=\langle \widehat{\Delta}_{(\myendogenousvals,\, \Omega)_\rho}, cc_{\lambda} \rangle$, where the set of probability measures $\widehat{\Delta}_{(\myendogenousvals,\, \Omega)_\rho} \subseteq \Delta_{(\myendogenousvals,\, \Omega)_\rho}$ is
\begin{equation}\label{eq:embedded_costalk_set}
    \widehat{\Delta}_{(\myendogenousvals,\, \Omega)_\rho} \coloneqq \{ \widehat{\chi}^\rho \text{ on } (\mathcal{V}, \Omega)_\rho \,:\, \alphamap{\myendogenous}^{\rho\trianglelefteq\tau}(\widehat{\chi}^\rho)=\chi^\tau \}\,,
\end{equation}
Please notice that \Cref{eq:embedded_costalk_set} and \Cref{th:ca_affine_functions} guarantee that $\widehat{E}(\scm{\rho})$ admits $E(\scm{\tau})$ as an IC CA.
We deliberately use the verbs \enquote{to project} and \enquote{to embed} to remark that in the node-edge transition formalized by the network sheaf, some of the causal knowledge idiosyncratic to the node is not transported into the backbone space; therefore, in the reverse edge-node transition formalized by the network cosheaf, it is only possible to integrate the transported causal knowledge into the causal knowledge of the node, with no guarantee of perfect reconstruction. 
This concept is essential for us to properly define the transfer of probability measures on the network.
At this point, we are ready to define the network sheaf and cosheaf of causal knowledge.
\begin{definition}[Network sheaf of causal knowledge]\label{def:net_sheaf_csprob}
    Given a network $G\coloneqq(\mathcal{N}, \mathcal{E})$ with face incidence poset $\faceincidenceposet$, a network sheaf valued in \CSprob is a functor $F: \faceincidenceposet \rightarrow \CSprob$ assigning (i) to each node $\rho$ and edge $\tau$ in \faceincidenceposet stalks consisting of convex spaces of probability measure, $E(\scm{\rho})$ and $E(\scm{\tau})$, respectively; (ii) to each node-edge incidence relation a restriction map $\alphamap{\myendogenous}^{\rho\trianglelefteq\tau}: E(\scm{\rho}) \rightarrow E(\scm{\tau})$ being the affine endogenous component within an IC \alphaabs between \scm{\rho} and \scm{\tau}.  
\end{definition}
\begin{definition}[Network cosheaf of causal knowledge]\label{def:net_cosheaf_csprob}
    Given a network $G\coloneqq(\mathcal{N}, \mathcal{E})$ with face incidence poset $\faceincidenceposet$, a network cosheaf valued in \CSprob is a functor $\widehat{F}: \faceincidenceposet^{\mathrm{op}} \rightarrow \CSprob$ assigning (i) to each node $\rho$ and edge $\tau$ in $\faceincidenceposet^{\mathrm{op}}$ costalks consisting of convex spaces of probability measure, $\widehat{E}(\scm{\rho})$ and $\widehat{E}(\scm{\tau})\equiv E(\scm{\tau})$, respectively; (ii) to each node-edge incidence relation an extension map $\betamap{\myendogenous}^{\rho\trianglelefteq\tau}: \widehat{E}(\scm{\tau}) \rightarrow \widehat{E}(\scm{\rho})$.  
\end{definition}
The network sheaf and cosheaf fulfill two complementary functions: the former transports causal knowledge from the nodes to a backbone space where such knowledge can be compared; the latter distributes causal knowledge from the backbone space to the network nodes.
\Cref{def:net_sheaf_csprob,def:net_cosheaf_csprob} are particular cases of more general network sheaf and cosheaf in \CSprob, in which every causal-related requirements on the (co)stalks and the (extension)restriction maps are dropped. For the network sheaf, we refer to a $0$-cochain as a collection of (interventional or observational) probability measures, one for each node.
The value of the $0$-cochain at the node $\rho$ is $\chi^\rho$.
Additionally, the $1$-cochain is a collection of probability measures on the edges, representing the IC CA of those on the nodes.
Hence, an induced value of the $1$-cochain at the  edge $\tau$ incident to a node $\rho$ is the pushforward probability measure $\chi^\tau=\alphamap{\myendogenous}^{\rho\trianglelefteq\tau}\left(\chi^\rho\right)$.
We denote the space of $0$- and $1$-cochains by $\mathcal{C}^0(G; F)$ and $\mathcal{C}^1(G; F)$, respectively.
An object of interest in sheaf theory \cite{curry2014sheaves} is the \emph{global section} of the network sheaf.
Loosely speaking, it is a consistent assignment of data to each node of the network that does not break local rules.
\begin{definition}[Global section]\label{def:global_section}
    Consider $F$ as in \Cref{def:net_sheaf_csprob}. A global section $\chi$ of $F$ is a choice $\chi^\rho$ in $E(\scm{\rho})$ for each node $\rho$ of $G$ such that $\chi^\tau=\alphamap{\myendogenous}^{\rho\trianglelefteq\tau}\left(\chi^\rho\right)$, for all $\rho\trianglelefteq\tau$.
    The space of global sections of $F$ is denoted by $\Gamma(G; F)$.
\end{definition}
Intuitively, in our RCK, a global section manifests itself as a collection of as many consistent (non-)interventions as the number of connected components of the network $G$. 
Specifically, consider a single connected component: according to our modeling, each pair of nodes connected by an edge admits a shared IC CA. 
Hence, choosing a consistent (non-)intervention for each node would result in a global section vanishing a certain notion of distance -- i.e., any suitable information-theoretic metric and $\phi$-divergence -- between the projected probability measures onto the edges. 
This is a direct consequence of the interventional consistency inherently ensured by the restriction maps.\\
Regarding the network cosheaf $\widehat{F}$, the $1$-chain agrees with the $1$-cochain of $F$, that is, a collection of probability measures representing the IC CA of those on the nodes.
The $0$-chain is a collection of probability measures, one for each node, satisfying the projection requirement in \Cref{eq:embedded_costalk_set}.
We denote by $\mathcal{C}^0(G; \widehat{F})$ and $\mathcal{C}^1(G; \widehat{F})$ the spaces of $0$- and $1$-chains of $\widehat{F}$.
Ultimately, we can pose the formal definition of \emph{relative causal knowledge}:
\begin{tcolorbox}[
    colback=white,
    colframe=black,
    boxrule=0.5pt,
    before=\par\smallskip\centering,
    after=\par\smallskip,
    boxsep=0pt
]
\begin{definition}[Relative Causal Knowledge]\label{def:rel_caus_know}
Given a network $G\coloneqq(\mathcal{N},\mathcal{E})$,  network sheaf and cosheaf of causal knowledge on $G$, $F$ and $\widehat{F}$, respectively, and two nodes $\rho_1$ and $\sigma_k$ connected by a path of $k$ edges $\tau_i:\rho_i \sim \sigma_i$, with $i \in [k]$, the relative causal knowledge is the causal knowledge at $\rho_1$ from the perspective of $\sigma_k$ filtered by the shared, abstract, causal knowledge on the edges $\tau_i$, $i \in [k]$ (i.e., path-dependent).
Specifically,
\begin{align}\label{eq:rel_caus_know}
        &\mathsf{CK}^{\rho_1,\sigma_k}\coloneqq\{\nonumber \\
        &\chi^{\rho_1,\sigma_k} \!=\! \betamap{\myendogenous}^{\sigma_k \trianglelefteq \tau_k} \!\circ\! \alphamap{\myendogenous}^{\rho_k \trianglelefteq \tau_k}\! \circ \! \ldots \! \circ \! \betamap{\myendogenous}^{\sigma_1 \trianglelefteq \tau_1} \!\circ \!\alphamap{\myendogenous}^{\rho_1 \trianglelefteq \tau_1} (\chi^\rho)\,,\nonumber\\
        & \,\chi^\rho \in E(\scm{\rho})\}\,.
\end{align}
\end{definition}
\end{tcolorbox}
We end the section with a working example on the transfer of causal knowledge.
Further discussion and exemplification of RCK is provided in \Cref{app:disc&ex}.

\spara{Setup.}
Consider a network with three nodes and two edges, forming a chain.
We assign to the leftmost node the causal knowledge of subject A, viz. $\mathsf{CK}^A$. Similarly, we attach to the central and rightmost nodes the causal knowledge of subject B and C, viz. $\mathsf{CK}^B$ and $\mathsf{CK}^C$.
The endogenous variables for the subjects are $\mathcal{A}=\{A_1,A_2,A_3\}$, $\mathcal{B}=\{B_1,B_2,B_3, B_4, B_5\}$, $\mathcal{C}=\{C_1,C_2,C_3\}$, composing non-intervened SMCs $\mathsf{M}^A$, $\mathsf{M}^B$, and $\mathsf{M}^C$.
Each subject does not have any information about the SCMs of the others.
$\mathsf{M}^A$ and $\mathsf{M}^B$ admit a shared CA $\mathsf{M}^{X}$ living on the shared edge, consisting of a single causal variable $X$ for simplicity. Specifically, for A the CA structure is $\{A_1, A_2\}\rightarrow X$; for B is $\{B_1,B_2,B_3\}\rightarrow X$.
$\mathsf{M}^B$ and $\mathsf{M}^C$ admit a shared CA $\mathsf{M}^{Y}$, given by a single causal variable $Y$. Here, for B the structure is $\{B_3,B_4,B_5\}\rightarrow Y$; for C is $\{C_1,C_2\}\rightarrow Y$.
For convenience, consider the restriction and extension maps, viz. $\alphamap{\myendogenous}^{\rho \trianglelefteq \tau}$ and $\betamap{\myendogenous}^{\rho \trianglelefteq \tau}$ to be linear (i.e., matrices). 
For A, we denote the restriction by $\mathbf{F}^{A\trianglelefteq X} \in \mathbb{R}^{1 \times 3}$ and the extension by $\widehat{\mathbf{F}}^{A\trianglelefteq X} \in \mathbb{R}^{3 \times 1}$. 
Similarly, for B we have $\mathbf{F}^{B\trianglelefteq X}$, $\mathbf{F}^{B\trianglelefteq Y}$ in $\mathbb{R}^{1 \times 5}$ as restrictions and $\widehat{\mathbf{F}}^{B\trianglelefteq X}$, $\widehat{\mathbf{F}}^{B\trianglelefteq Y}$ in $\mathbb{R}^{5 \times 1}$ as extensions.
Finally, for C we have $\mathbf{F}^{C\trianglelefteq X} \in \mathbb{R}^{1 \times 3}$ and $\widehat{\mathbf{F}}^{C\trianglelefteq X} \in \mathbb{R}^{3 \times 1}$.
Zero entries in the restriction and extension matrices manage the fact that some variables are not relevant to CA, a.k.a. non-constructive abstraction.

\spara{Relative Causal Knowledge.}
The subjects perform certain tasks through individual soft-intervened SCMs,
entailing the soft-intervened probability measures $\chi_\mathcal{S}^A=N(0, \boldsymbol{\Sigma}_\mathcal{S}^A)$, $\chi_\mathcal{S}^B=N(0, \boldsymbol{\Sigma}_\mathcal{S}^B)$, and $\chi_\mathcal{S}^C=N(0, \boldsymbol{\Sigma}_\mathcal{S}^C)$. 
The collection of these measures constitutes a valuation of the node stalks of the network sheaf, i.e., a $0$-cochain.
When projected onto the edges via the above CAs, the $0$-cochain entails the $1$-cochain, that is, a collection of probability measures for the edges $X$ and $Y$.
Then, \Cref{def:rel_caus_know} specifies for instance how B and C see A (i.e., $\chi_\mathcal{S}^A$) from their perspectives.
Specifically, the relative soft-intervened measures are:\\
\emph{(i)} for B, $\chi_\mathcal{S}^{A,B}=N(0,\widehat{\mathbf{F}}^{B\trianglelefteq X} \mathbf{F}^{A\trianglelefteq X}\boldsymbol{\Sigma}_\mathcal{S}^A \mathbf{F}^{{A\trianglelefteq X}^\top}\widehat{\mathbf{F}}^{{B\trianglelefteq X}^\top})$ $=N(0, \boldsymbol{\Sigma}_\mathcal{S}^{A,B})$;\\
\emph{(ii)} for C, $\chi_\mathcal{S}^{A,C}=N(0,\widehat{\mathbf{F}}^{C\trianglelefteq Y} \mathbf{F}^{B\trianglelefteq Y}\boldsymbol{\Sigma}_\mathcal{S}^{A,B}\mathbf{F}^{{B\trianglelefteq Y}^\top}\widehat{\mathbf{F}}^{{C\trianglelefteq Y}^\top})$ $=N(0, \boldsymbol{\Sigma}_\mathcal{S}^{A,C})$.

\spara{Global section.}
A key point of our framework is the possibility of investigating the global agreement among subjects in terms of causal knowledge. 
Such a global agreement gives rise to the global section in \Cref{def:global_section}.
In the working example, we have a global section when the soft-interventions performed by the subjects are such that \\
$\mathbf{F}^{A\trianglelefteq X}\boldsymbol{\Sigma}_\mathcal{S}^A \mathbf{F}^{{A\trianglelefteq X}^\top}=\mathbf{F}^{B\trianglelefteq X}\boldsymbol{\Sigma}_\mathcal{S}^B \mathbf{F}^{{B\trianglelefteq X}^\top}$ (section on $X$)\\
and\\
$\mathbf{F}^{B\trianglelefteq Y}\boldsymbol{\Sigma}_\mathcal{S}^{B}\mathbf{F}^{{B\trianglelefteq Y}^\top}=\mathbf{F}^{C\trianglelefteq Y}\boldsymbol{\Sigma}_\mathcal{S}^{C}\mathbf{F}^{{C\trianglelefteq Y}^\top}$ (section on $Y$).

%% file: sections/discussion.tex
% !TEX root =  ../main.tex
\section{Discussion}\label{sec:discussion}

\noindent Conceptually, the relativity of causal knowledge can drive a paradigm shift in how causality is typically understood in AI/ML. 
By stripping causality of its oracular and absolute meaning, the relativity of causal knowledge situates it within a different ontological setting, where truth is not monolithic but emerges inevitably and relatively from a set of relationships. 
This is mathematically formalized by the path-dependent RCK in \Cref{eq:rel_caus_know}.
Our framework paves the way to multiple intriguing areas of reasearch, pivotal to fully characterize RCK and make it applicable. \\
\noindent\textbf{Learning Theory.} 
We developed our theoretical framework with an underlying intuition: RCK could be a better framework to empower AI/ML with the ladder of causation \cite{pearl2018book}. 
A learning theory is needed for this. 
It is surely of interest to develop a statistical theory for models that can ingest network sheaves and cosheaves of causal knowledge to solve downstream learning tasks. 
In many cases, the restriction and expansion maps -- the formal instance of \enquote{perspective} -- are given, e.g., privacy requirements of the subjects in an agentic network or communication constraints in a sensor network. 
However, when they are not available, the problem of \emph{(co)sheaf inference}, i.e., learning them in a data-driven fashion given the network connectivity, is fundamental. 
An even harder problem is \emph{(co)sheaf discovery}, i.e., jointly learning the restriction, extension maps, and network connectivity. 
Interestingly, both for inference and discovery, the learning losses and models are useful to induce a particular notion of perspective, that could be fairness-, safety-, information-, or trustworthiness-driven, just to name a few.\\
\noindent\textbf{Cohomology Theory.} 
\CSprob is not an Abelian category, as it is not additive, i.e.,  there is no natural way to \enquote{add} two affine maps to get another affine map such that all group axioms are globally satisfied. 
Consequently, defining a cochain complex -- and thus a cohomology theory -- on network sheaves of causal knowledge is nontrivial.
Developing such a theory would yield rich algebraic invariants that offer global insights into the network structure. 
These invariants would be instrumental in revealing when and why multiple sheaves of CK share a common structure.
Practically, a cohomology theory could help us understand how and why different networks of subjects, each developing causal knowledge about different phenomena, exhibit similarities.\\
\noindent\textbf{Hodge Theory.} 
The Hodge theorem \cite{griffiths2014principles} shows that the kernel of the Laplacian on a Riemannian manifold recovers its de Rham cohomology.  
Combinatorial versions of the Hodge theory have been developed for cellular sheaves of inner-product spaces (e.g., vector spaces and Hilbert spaces), building on suitable properties of the adjoint maps \cite{hansen2019toward}. 
Unfortunately, porting these arguments to the $\CSprob$ category is nontrivial.
In this context, a Hodge(-like) theory could be instrumental in defining diffusion-type operators and developing spectral theories \cite{hansen2019toward} for RCK. 
Practically, having a Hodge theory might help us understand which are the dynamics that lead to global sections in \Cref{def:global_section} and why, providing insights on when and why RCK aligns.

%% file: sections/conclusion.tex
% !TEX root =  ../main.tex
\section{Conclusion}\label{sec:conclusion}
In this paper, we introduced the \emph{relativity of causal knowledge}.
We established the formal definition of \emph{relative causal knowledge} (RCK) via a category-theoretic treatment for SCMs, their interventions, and related $\boldsymbol{\alpha}$-abstractions.
Within the relativity of the causal knowledge paradigm, SCMs are inherently imperfect, subjective representations embedded within networks of relationships.  
By encoding causal knowledge in convex spaces of probability measures and constructing network sheaves and cosheaves, we enabled the transfer of causal knowledge in an interventionally consistent manner while considering diverse perspectives across the network.
Our work opens exciting research avenues in cohomology, Hodge(-like), and learning theory for RCK, paving the way for context-aware, robust AI systems.

%% file: appendix/background.tex
% !TEX root =  ../main.tex
\section{Essential material on category theory}\label{app:background}
Below are fundamental definitions and examples that are instrumental in providing the necessary background on category theory.
For a comprehensive overview of category theory see resources such as \cite{mac2013categories,perrone2024starting}.

\begin{definition}[Category]\label{def:category}
    A category $\mathsf{C}$ consists of
    \begin{squishlist}
        \item A collection $\mathcal{C}_0$ whose elements are called objects of $\mathsf{C}$;
        \item A collection $\mathcal{C}_1$ whose elements are called morphisms of $\mathsf{C}$;
    \end{squishlist}
    such that:
    \begin{squishlist}
        \item Each morphism $f : X \rightarrow Y$ has assigned two objects of the category called source and target, respectively;
        \item Each object $X$ has a distinguished identity morphism $\catidentity_X: X \rightarrow X$;
        \item Given $f: X\rightarrow Y $ and $g:Y\rightarrow Z$, than it exists $g \circ f = h: X \rightarrow Z$.
    \end{squishlist}
    These structures satisfy the following axioms:
    \begin{squishlist}
        \item (Unitality) $\forall f: X \rightarrow Y, \; f \circ \catidentity_X=f \text{ and } \catidentity_Y \circ f = f$;
        \item (Associativity) Given $f$, $g$, and $h$ such that the compositions are defined, then $h \circ (g \circ f) = (h \circ g) \circ f$.
    \end{squishlist}
\end{definition}

\begin{example}
    The following are some notable examples of categories:
    \begin{squishlist}
        \item Indicate with \Poset a partial order set. \Poset can be viewed as the category whose objects are the elements $p \in \mathcal{P}$ and morphisms are order relations $p \leq p^\prime$. Notice that there is at most one morphism between two objects;
        \item \Vect is the category whose objects are real vector spaces and morphisms are linear maps;
        \item \Hilb is the category whose objects are real complete Hilbert spaces, that is, real vector spaces equipped with an inner product structure and closed under the norm topology.
        Morphisms are bounded linear maps $f: X \rightarrow Y$, that is, $\exists \, b \in \reall^+$ such that $\norm{f(x)} \leq b \norm{x}, \; \forall x \in X$. 
        Every morphism $f: X \rightarrow Y$ between Hilbert spaces admits an \emph{adjoint} $f^\star$ such that, $\forall \, x \in X$ and $y \in Y$, $\Eprod{y}{f(x)}{}=\Eprod{f^\star(y)}{x}{}$.
        We also have $(f^\star)^\star=f$.
    \end{squishlist}
\end{example}

Starting from \Cref{def:category} we can define the notion of \emph{subcategory}.

\begin{definition}\label{def:subcategory}
    Given a category $\mathsf{C}$, a subcategory $\mathsf{S}$ of $\mathsf{C}$ consists of the following data:
    \begin{squishlist}
        \item A subcollection $\mathcal{S}_0$ of $\mathcal{C}_0$ such that $\forall \, S, \, \catidentity_S$ is in $\mathsf{S}$;
        \item A subcollection $\mathcal{S}_1$ of $\mathcal{C}_1$ such that $\forall \, f \text{ and } g$ for which the composition is defined in $\mathsf{C}$, the composite $h = g \circ f$ is in $\mathsf{S}$.
    \end{squishlist}
\end{definition}

\begin{example}[\IPVect]\label{ex:IPVect}
The category of real vector spaces with an inner product \IPVect is a subcategory of \Vect. 
Specifically, objects are real vector spaces equipped with an inner product structure.
Morphisms are bounded linear maps $f: X \rightarrow Y$, that is, $\exists \, b \in \reall^+$ such that $\norm{f(x)} \leq b \norm{x}, \; \forall x \in X$. 
Every morphism $f: X \rightarrow Y$ between inner product vector spaces admits an \emph{adjoint} $f^\star$ such that, $\forall \, x \in X$ and $y \in Y$, $\Eprod{y}{f(x)}{}=\Eprod{f^\star(y)}{x}{}$.
We also have $(f^\star)^\star=f$.
\IPVect is similar to \Hilb, the difference is that in \Hilb the vector spaces are also complete.
\end{example}

\begin{definition}[Functor]\label{def:functor}
    Consider $\mathsf{C}$ and $\mathsf{D}$ categories. 
    A functor $F: \mathsf{C} \rightarrow \mathsf{D}$ consists of the following data:
    \begin{squishlist}
        \item For each object $X \in \mathcal{C}_0$, an object $F(X) \in \mathcal{D}_0$;
        \item For each object morphism $\mathcal{C}_1 \ni f: X \rightarrow Y$, a morphism $\mathcal{D}_1 \ni F(f): F(X) \rightarrow F(Y)$;
    \end{squishlist}
    such that the following axioms hold:
    \begin{squishlist}
        \item (Unitality) $\forall X \in \mathcal{C}_0, \; F(\catidentity_X) \!=\! \catidentity_F(X)$. In other words, the identity in $\mathsf{C}$ is mapped into the identity in $\mathsf{D}$.
        \item (Compositionality) $\forall f \text{ and } g \in \mathcal{C}_1$ such that the composition is defined, then $F(g \circ f) = F(g) \circ F(f)$. In other words, the composition in $\mathsf{C}$ is mapped into the composition in $\mathsf{D}$.
    \end{squishlist}
\end{definition}

To ease the notation, in the sequel, we use $F^\sigma$ and $F^f$ to denote $F(\sigma)$ and $F(f)$, respectively.

\begin{definition}[Natural transformation]\label{def:nat_transf}
Consider two categories $\mathsf{C}$ and $\mathsf{D}$, and two functors between them, namely $F: \mathsf{C} \rightarrow \mathsf{D}$ and $G: \mathsf{C} \rightarrow \mathsf{D}$.
A natural transformation $\alpha: F \rightarrow G$ consists of the following data:
\begin{squishlist}
    \item For each object $X \in \mathcal{C}_0$, a morphism $\alpha_{X}: F^X \rightarrow G^X$ in $\mathcal{D}$ called the component of $\alpha$ at $X$;
    \item For each morphism $f: X \rightarrow X^\prime$ in $\mathcal{C}$, the following diagram commutes:
    \begin{equation}
        \begin{tikzcd}[row sep=1.5cm, column sep=1.5cm]
            F^X \arrow[r, "F^f"] \arrow[d, "\alpha_X"'] & F^{X^\prime} \arrow[d, "\alpha_{X^\prime}"] \\
            G^X \arrow[r, "G^f"'] & G^{X^\prime}
        \end{tikzcd}
    \end{equation}
\end{squishlist}
\end{definition}

A natural transformation can be thought of as a consistent system of arrows between two functors.
The \emph{naturality} of $\alpha$ means that it is invariant with respect to maps between the images of two functors.

%% file: appendix/examples.tex
% !TEX root =  ../main.tex
\section{Examples}\label{app:examples}

This appendix is devoted to provide practical examples about the formalism introduced in \Cref{sec:cat_scm_ck}.
Consider $\myexogenous=\{Z_1, Z_2, Z_3\}$ and $\myendogenous=\{X_1, X_2, X_3\}$ as exogenous and endogenous variables of the following linear SCM with additive noise $\mathsf{M}$:
\begin{equation}\label{eq:linSCM_noni}
\begin{cases}
    X_1 = Z_1 = m_1(Z_1)\, ,\\
    X_2 = c_{21}X_1 + Z_2 = c_{21}Z_1 + Z_2 = m_2(Z_1,Z_2)\, ,\\
    X_3 = c_{32} X_2 + Z_3 = c_{32}c_{21}Z_1 + c_{32}Z_2 +Z_3 = m_3(Z_1,Z_2,Z_3)\,;
\end{cases}    
\end{equation}
where $Z_i \sim N(0,1)$ and $Z_i \perp Z_j$, $i, j \in [3]$ and $i \neq j$.
Additionally, let the probability space associated with the exogenous to be $(\reall^n, \mathcal{B}(\reall^n), N(\zeros_3, \identity_3) )$, where $\mathcal{B}(\reall^n)$ is the Borel $\sigma$-algebra of $\reall^n$.
We can equivalently express $\mathsf{M}$ as $\X=\mixing\Z$, where
\begin{equation}
    \mixing = \begin{bmatrix}
    1 & 0 & 0\\
    c_{21} & 1 & 0\\
    c_{32}c_{21} & c_{32} & 1
    \end{bmatrix}\,.    
\end{equation}
We call the matrix $\mixing$ the mixing matrix.
See Lemma E.1 in \citet{d'acunto2023learning} for a proof of the existence of $\mixing$.

At this point, according to \Cref{def:SCM_meas}, the components of the measure-theoretic \scm{} are:
\emph{(i)} $(\reall^n, \mathcal{B}(\reall^n), N(\zeros_3, \identity_3) )$ for the exogenous;
\emph{(ii)} $\mixing$ as mixing;
and \emph{(iii)} $(\reall^n, \mathcal{B}(\reall^n), N(\zeros, \mixing\mixing^\top))$ for the endogenous.
In detail, the endogenous measure is the pushforward of $N(\zeros, \identity_3)$ given \mixing. 
Using the notation in the paper for pushforward measures, each $X_i$ follows $m_i(\mu(Z_i \cup \myexogenous^{\ancestors_i}))=m_i(N(\zeros,\identity_{|\ancestors_i|+1}))=N(0, [\mixing]_{i,:}[\mixing]_{i,:}^\top)$.

Starting from the measure-theoretic SCM, according to \Cref{def:scm_fun}, the category-theoretic representation of \scm{} is the functor mapping $I$ to $(\reall^n, \mathcal{B}(\reall^n), N(\zeros_3, \identity_3) )$, $I^\prime$ to $(\reall^n, \mathcal{B}(\reall^n), N(\zeros, \mixing\mixing^\top))$, and the arrow between $I$ and $I^\prime$ to \mixing.

At this point, let us consider hard and soft interventions.
Starting with the former, we consider the hard intervention $\mathrm{do}(X_2=c)$.
Accordingly, \Cref{eq:linSCM_noni} becomes
\begin{equation}\label{eq:linSCM_hi}
    \begin{cases}
        X_1 = Z_1 = m_1^\hard(Z_1) \sim N(0,1)\, ,\\
        X_2 = c \quad \text{(constant),} \\
        X_3 = c_{32} c + Z_3 = d + Z_3 = m^\hard_3(Z_3) \sim N(d,1)\,.
    \end{cases}  
\end{equation}
Starting from \Cref{eq:linSCM_hi}, the hard-intervened SCM \scm{\hard} in \SCMcat is the functor mapping $I$ to $(\reall^n, \mathcal{B}(\reall^n), N(\zeros_3, \identity_3) )$, $I^\prime$ to $(\reall^n, \mathcal{B}(\reall^n), N(0, 1) \times \delta_c \times N(d,1))$, and the arrow between $I$ and $I^\prime$ to $\mymixing_\hard=\{m_1^\hard, c ,m_3^\hard\}$. 

Consider now a soft intervention on $X_2$ which modifies $c_{21} \to c_{21}^\soft$.
Recall that we do not consider soft interventions that modify the parent set of the intervened variable.
Accordingly, \Cref{eq:linSCM_noni} becomes
\begin{equation}\label{eq:linSCM_si}
    \begin{cases}
        X_1 = Z_1 = m^\soft_1(Z_1)\, ,\\
        X_2 = c_{21}^\soft X_1 + Z_2 = c_{21}^\soft Z_1 + Z_2 = m^\soft_2(Z_1,Z_2)\, ,\\
        X_3 = c_{32} X_2 + Z_3 = c_{32}c_{21}^\soft Z_1 + c_{32}Z_2 +Z_3 = m^\soft_3(Z_1,Z_2,Z_3)\,;
    \end{cases}
\end{equation}
For convenience, represent $\mymixing_\soft$ as
\begin{equation}
    \mixing_\soft = \begin{bmatrix}
    1 & 0 & 0\\
    c_{21}^\soft & 1 & 0\\
    c_{32}c_{21}^\soft & c_{32} & 1
    \end{bmatrix}\,.    
\end{equation}
At this point, the soft-intervened SCM \scm{\soft} in \SCMcat is the functor mapping $I$ to $(\reall^n, \mathcal{B}(\reall^n), N(\zeros_3, \identity_3) )$, $I^\prime$ to $(\reall^n, \mathcal{B}(\reall^n), N(\zeros, \mixing_\soft \mixing_\soft^\top) )$, and the arrow between $I$ and $I^\prime$ to $\mixing_\soft$.

Finally, we exemplify the category-theoretic representation of the above interventions in \SCMcat.
The inverse $\mymixing^{-1}$ exists by definition of the mixing matrix \mixing itself.
Specifically, it is equal to $\identity_3 - \mathbf{C}$, where $\mathbf{C}$ is the matrix of causal coefficients corresponding to \Cref{eq:linSCM_noni}.
Using \Cref{lem:intervention_prob}, 
in \SCMcat \emph{(i)} the hard intervention corresponds to the natural transformation $\eta^\hard=\langle \catidentity_{(\reall^n, \mathcal{B}(\reall^n), N(\zeros_3, \identity_3) )}, \hard = \mymixing_\hard \circ \mymixing^{-1}\rangle$;
and \emph{(ii)} the soft intervention corresponds to the natural transformation $\eta^\soft=\langle \catidentity_{(\reall^n, \mathcal{B}(\reall^n), N(\zeros_3, \identity_3))}, \soft = \mymixing_\soft \circ \mymixing^{-1}\rangle$.

%% file: appendix/invertibility.tex
% !TEX root =  ../main.tex
\section{On the invertibility assumption}\label{app:invertibility}
Here we illustrate that invertibility of \mymixing holds for the main classes of Markovian SCMs.

\spara{Additive Noise Model.}
In this case, considering $n$ causal variables, the structural equations read as
\begin{equation}\label{eq:anm}
    X_i = f_i(\parents_i) + Z_i, \quad \forall\, i \in [n]\,.
\end{equation}
It is immediate to obtain that the components of $\mymixing^{-1}: \myendogenousvals \rightarrow \myexogenousvals$ are $m^{-1}_i = X_i - f_i(\parents_i) $, $\forall\, i \in [n]$.

\spara{Post-nonlinear Model.}
In this case, the structural equations read as
\begin{equation}\label{eq:pnl}
    X_i = g_i(f_i(\parents_i) + Z_i), \quad \forall\, i \in [n]\,;
\end{equation}
where $g_i$ is invertible \cite{zhang2009identifiability,kaltenpoth2023nonlinear}.
It is immediate to obtain that the components of $\mymixing^{-1}: \myendogenousvals \rightarrow \myexogenousvals$ are $m^{-1}_i = g_i^{-1}(X_i) - f_i(\parents_i) $, $\forall\, i \in [n]$.

\spara{Location-scale Noise Models.}
In this case, the structural equations are \cite{immer2023identifiability}
\begin{equation}\label{eq:lsnm}
    X_i = f_i(\parents_i) + h_i(\parents_i)Z_i, \quad \forall\, i \in [n]\,;
\end{equation}
where $f_i: \myendogenousvals_{\parents_i} \rightarrow \reall$ and $h_i: \myendogenousvals_{\parents_i} \rightarrow \reall_+$ is strictly positive.
It is immediate to obtain that the components of $\mymixing^{-1}: \myendogenousvals \rightarrow \myexogenousvals$ are $m^{-1}_i = \left(X_i - f_i(\parents_i) \right)/h_i(\parents_i)$, $\forall\, i \in [n]$.

%% file: appendix/proofs.tex
% !TEX root =  ../main.tex
\section{Proofs}\label{app:proofs}

\interventionlemma*
\begin{equation}
    \begin{tikzcd}[row sep=1.5cm, column sep=1.5cm]
        \scm{}(I) \arrow[r, "\mymixing"] \arrow[d, "\catidentity_{\scm{}(I)}"'] & \scm{}(I^\prime) \arrow[d, "\intervention"] \\
        \scm{}_{\intervention}(I) \arrow[r, "\mymixing_\intervention"'] & \scm{}_\mathcal{H}({I^\prime)}
    \end{tikzcd} \Longrightarrow  \mymixing_\intervention = \intervention \circ \mymixing\,. 
    \tag{(7)}
\end{equation}
\begin{proof}
    Let us start from the exogenous component.
    Since the intervention \intervention, that could be either soft or hard, acts only on the endogenous variables, the probability space corresponding to the exogenous part remains unchanged. 
    That is, for the object $\scm{}(I)$ we simply have
    \begin{equation}
        \eta^\intervention_I \coloneqq \catidentity_{\scm{}(I)}\,.
    \end{equation}
    This means that for any measure $\mu$ on the exogenous, the pushforward along $\eta^\intervention_I$ is exactly $\mu$.

    Now, let us consider the endogenous component and recall the collections $\mathcal{F}_\hard$ and $\mathcal{F}_\soft$ in \Cref{sec:cat_scm_ck}. 
    If the intervention is hard, than we represent it as the collection $\mathcal{F}_\intervention \coloneqq \mathcal{F}_\hard$.
    If soft, with $\mathcal{F}_\intervention \coloneqq \mathcal{F}_\soft$.
    The collection $\mathcal{F}_\intervention$ entails a new collection of mixing functions; specifically \emph{(i)} $\mymixing_\hard$ for the hard intervention case, \emph{(ii)} $\mymixing_\soft$ for the soft one (cf. \Cref{sec:cat_scm_ck}).
    At this point, by invertibility of \mymixing, \Cref{eq:nat_transf_inter_SCMcat} trivially follows by setting the second component $\eta^\intervention_{I^\prime}$ equal to $\intervention=\mymixing_\intervention \circ \mymixing^{-1}$.
\end{proof}

\convexspaceprob*
\begin{proof}
    According to Def. 3.1 in \cite{fritz2009convex}, we have to demonstrate that $cc_{\lambda}(\chi_1,\chi_2)$ satisfies: \emph{(i)} $cc_0(\chi_1,\chi_2)=\chi_2$, \emph{(ii)} $cc_0(\chi_1,\chi_1)=\chi_1$; \emph{(iii)} $cc_{\lambda}(\chi_1,\chi_2)=cc_{\bar{\lambda}}(\chi_2,\chi_1)$; and $cc_{\lambda}(cc_{\mu}(\chi_1,\chi_2),\chi_3)=cc_{\widetilde{\lambda}}(\chi_1, cc_{\widetilde{\mu}}(\chi_2,\chi_3))$ where:
    \begin{equation}
        \widetilde{\lambda}=\lambda \mu, \quad \quad \quad \widetilde{\mu} = \begin{cases}
            \frac{\lambda \bar{\mu}}{\bar{\lambda \mu}}, \quad \text{if } \lambda\mu \neq 1\,,\\
            \text{arbitrary}, \quad \text{if } \lambda=\mu=1\,.
        \end{cases}
    \end{equation}
    Since the convex set $\Delta_{(\myendogenousvals,\Omega)}$ is a subset of a vector space, the previous properties follow from the axioms of vector space.
\end{proof}

\csprobcat*
\begin{proof}    
    From Definition 3.2 in \cite{fritz2009convex} we have that convex spaces as in \Cref{lem:convexspace_prob} together with morphisms commuting with \(cc_\lambda\) form a category. Hence we have to demonstrate that affine measurable maps commute with $cc_\lambda$.
    Recall that a map $f: \Delta_{(\myendogenousvals,\, \Omega)} \to \Delta_{(\myendogenousvals^\prime,\, \Omega^\prime)}$
    is affine if for every $\chi_1,\,\chi_2 \in \Delta_{(\myendogenousvals,\, \Omega)}$ and every $\lambda \in [0,1]$ it satisfies
    $f\left(cc_\lambda(\chi_1,\chi_2)\right) = cc_\lambda\left(f(\chi_1),f(\chi_2)\right).$
    Hence, by definition, any affine measurable map necessarily commutes with $cc_\lambda$. 
\end{proof}

\convexcomb*
\begin{proof}
    Consider $\scm{}=\langle (\myexogenousvals,\, \Upsilon, \zeta), \, (\myendogenousvals,\, \Omega, \chi)\, , \mymixing \rangle$, $\lambda \in [0,1]$, and w.l.o.g. the probability measures $\chi_{\mathcal{S}_1}$ and $\chi_{\mathcal{S}_2}$ entailed by two (measurable) soft interventions $\eta^{{S}_1}$ and $\eta^{\mathcal{S}_2}$ run on \scm{}.
    Hence, the convex combination
    \begin{equation}
        \chi_{\mathcal{S}_3} = cc_{\lambda}(\chi_{\mathcal{S}_1}, \chi_{\mathcal{S}_2}) = \lambda \chi_{\mathcal{S}_1} + \bar{\lambda} \chi_{\mathcal{S}_2}
    \end{equation}
    is a proper soft-interventional probability measure entailed by $\mymixing_{\soft_3}=\{m^{\soft_3}_1, \ldots, m^{\soft_3}_n\}$ according to \Cref{eq:scm_sint_push}, where $m^{\soft_3}_i=\lambda m^{\soft_1}_i + \bar{\lambda} m^{\soft_2}_i$.
    Indeed we have
    \begin{equation}
        \begin{aligned}
            \lambda \chi_{\mathcal{S}_1} + \bar{\lambda} \chi_{\mathcal{S}_2} &=\bigtimes_{X_i\in \myendogenous} \Big( \lambda m^{\soft_1}_i\left(\mu_i \left( \myexogenousvals_i \times \myexogenousvals^{\ancestors_i} \right) \right) +\\
            & \quad \quad \quad \quad + \bar{\lambda} m^{\soft_2}_i\left(\mu_i \left( \myexogenousvals_i \times \myexogenousvals^{\ancestors_i} \right) \right) \Big)\\
            &=\bigtimes_{X_i\in \myendogenous} \left(\lambda m^{\soft_1}_i + \bar{\lambda} m^{\soft_2}_i\right)\left(\mu_i \left( \myexogenousvals_i \times \myexogenousvals^{\ancestors_i} \right) \right)\\
            &= \bigtimes_{X_i\in \myendogenous} m^{\soft_3}_i\left(\mu_i \left( \myexogenousvals_i \times \myexogenousvals^{\ancestors_i} \right) \right) = \chi_{\mathcal{S}_3}\,.\\
        \end{aligned}
    \end{equation}
    Also, recall that a convex combination of measurable functions is still measurable.
    Starting from \Cref{eq:scm_obs_push,eq:scm_hint_push,eq:nat_transf_inter_SCMcat}, it is straightforward to see that the same holds also when we consider observational and hard-interventional measures within the convex combination.
\end{proof}

\caaffinefunctions*
\begin{proof}
    Consider two SCMs \scm{} and $\scm{\prime}$ related by an IC \alphaabs as in \cref{def:IC_alpha_abstraction_scmcat}, and interventions $\mathcal{I}_1^\prime$ and $\mathcal{I}_2\prime$ on $\scm{\prime}$, either hard or soft, corresponding to interventions $\mathcal{I}_1$ and $\mathcal{I}_2$ on $\scm{}$.
    Then, given $\intervention_3^\prime$ convex combination of $\mathcal{I}_1^\prime$ and $\mathcal{I}_2\prime$, we have
    \begin{equation}\label{eq:proof_alpha_affine}
        \begin{aligned}
            \chi'_{\intervention_3^\prime} &= \alphamap{\myendogenous}\left(\chi_{\intervention_3}\right)\\
            &\stackrel{(a)}{=} \alphamap{\myendogenous}\left(\lambda \chi_{\intervention_1}+ \bar{\lambda} \chi_{\intervention_2}\right) \\
            &\stackrel{(b)}{=} \lambda \chi'_{\intervention_1^\prime}+ \bar{\lambda} \chi'_{\intervention_2^\prime} = \chi'_{\intervention_3^\prime}\,.
        \end{aligned}
    \end{equation}
    where in \emph{(a)} we apply \Cref{th:convex_comb_prob_meas}, and in \emph{(b)} the linearity of the pushforward.
\end{proof}

\encodingfunctor*
\begin{proof}
    The mapping on objects is well-defined since \Cref{th:convex_comb_prob_meas} guarantees that convex combinations of probability measures associated with observational and interventional states of \scm{} are valid soft-interventional probability measures.
    On morphisms, \Cref{th:ca_affine_functions} implies that \alphamap{\myendogenous} is a proper morphism in \CSprob.
    Interventional consistency is inherited because in \NI every morphism is an IC \alphaabs.
\end{proof}

%% file: appendix/disc_ex.tex
% !TEX root =  ../main.tex
\section{Further Discussion and Exemplification of RCK}\label{app:disc&ex}

\begin{figure*}
    \centering
\includegraphics[width=.9\linewidth]{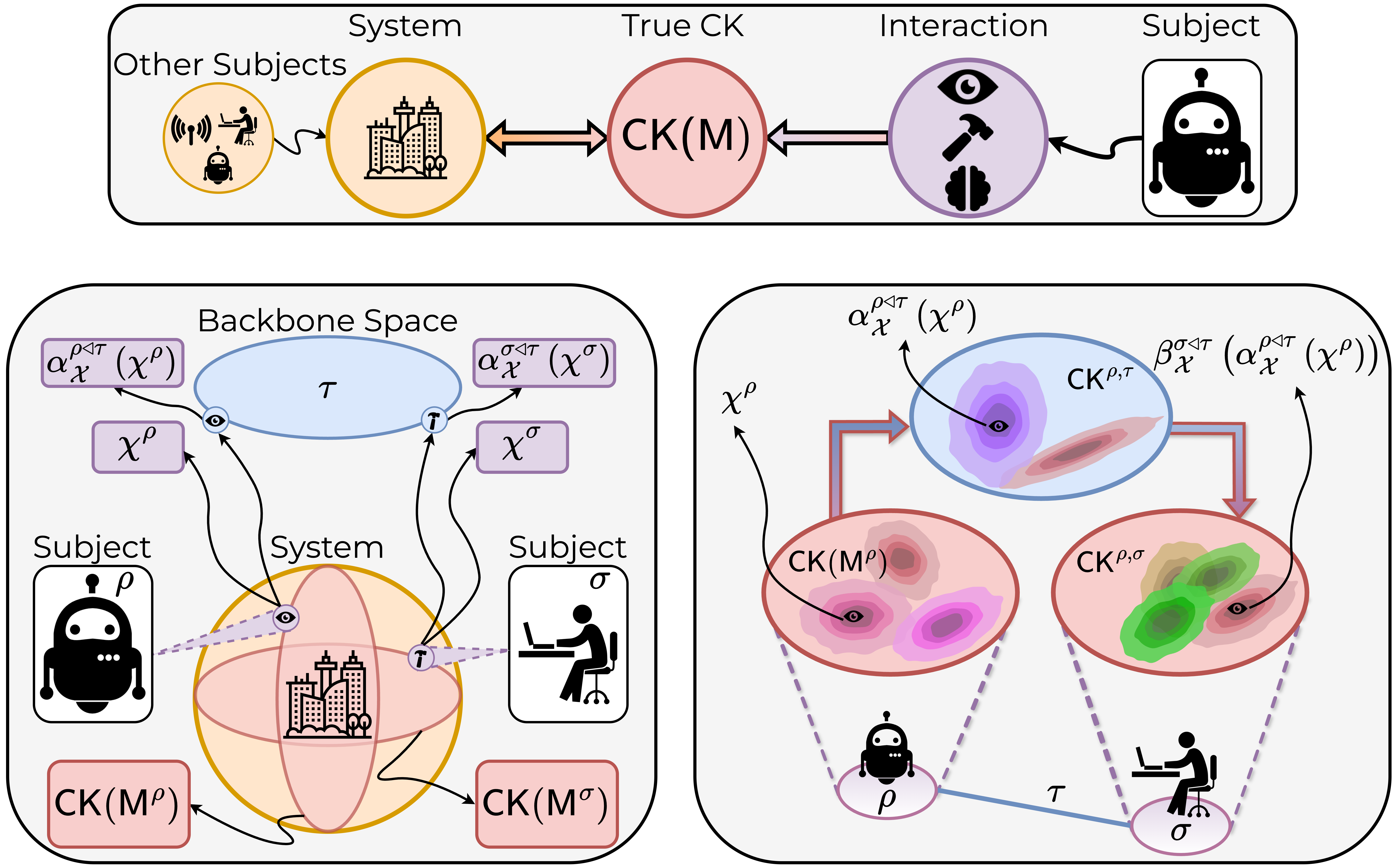}
    \caption{In the top figure, we depict the classical, non-relative approach to causality: a subject interacts with a \textcolor{system}{system} that is assumed to be completely describable by an underlying \enquote{true} SCM $\mathsf{M}$. We refer to the (convex) space of all the interventional and observational probability measures entailed by $\mathsf{M}$ as the \enquote{true} \textcolor{ck}{Causal Knowledge $\mathsf{CK}(\mathsf{M})$}. In this classical setting, the \textcolor{system}{system} and $\mathsf{M}$ are thus in a bijection, the other subjects are assumed to be part of the \textcolor{system}{system}, and the observer subject interacts with the \textcolor{system}{system} following (or inferring) \textcolor{ck}{$\mathsf{CK}(\mathsf{M})$}. The \textit{Relativity of Causal Knowledge} fundamentally challenges this paradigm by making causality subjective and relational, i.e., by breaking the bijection between the \textcolor{system}{system} and a \enquote{true} SCM, and isolating the subjects from the \textcolor{system}{system} while allowing them to interact with each other. The novel, core mathematical object implementing our relative paradigm is the \textit{Network (Co)Sheaf of Causal Knowledge}, which can be informally thought of as a graph on whose nodes and edges are attached certain convex spaces, called \textit{stalks}, that can interact through certain mappings, called \textit{restriction and extension maps}. In our relative setting, as we show in the bottom left figure, the \textcolor{system}{system} can be pictured as a spherical planet, and the subjects as satellites orbiting around it. Each subject can only observe the planet from certain angles--its \textit{perspective}--, positioning it on a specific orbit. The orbits represent then the subjective \textcolor{ck}{Causal Knowledge} of the subjects--\textcolor{ck}{$\mathsf{CK}(\mathsf{M}^{\rho})$} and \textcolor{ck}{$\mathsf{CK}(\mathsf{M}^{\sigma})$}--, i.e., the (convex) spaces of all the interventional and observational probability measures entailed by the subjective SCMs that the subjects use to describe the underlying \textcolor{system}{system}--$\mathsf{M}^{\rho}$ and $\mathsf{M}^{\sigma}$. In sheaf jargon, each subject is a node, and each space \textcolor{ck}{$\mathsf{CK}(\mathsf{M}^{\cdot})$} is a \textcolor{ck}{\textit{node stalk}}. As such, a point \textcolor{cochain}{$\chi^{\rho}\in \mathsf{CK}(\mathsf{M}^{\rho})$} on the orbit of a subject $\rho$ is a \textit{specific probability measure}. In sheaf jargon, a collection of points, one per each subject's orbit --\textcolor{cochain}{$\{\chi^{\rho}, \chi^{\sigma}\}$}-- is a \textcolor{cochain}{\textit{0-cochain}}. Two subjects $\rho$ and $\sigma$ can communicate if their SCMs $\mathsf{M}^{\rho}$ and $\mathsf{M}^{\sigma}$ admit \textit{a shared abstraction}, i.e., a backbone space \textcolor{backbone}{$\tau$}. In this case, $\rho$ and $\sigma$ can map their 0-cochain values $\chi^{\rho}$ and $\chi^{\sigma}$ to more abstract  representations in the backbone space \textcolor{backbone}{$\alpha_{\mathcal{X}}^{\rho \triangleleft \tau}\left(\chi^\rho\right)$} and \textcolor{backbone}{$\alpha_{\mathcal{X}}^{\sigma \triangleleft \tau}\left(\chi^\sigma\right)$} through surjective mappings $\alpha_{\mathcal{X}}^{\rho \triangleleft \tau}$ and $\alpha_{\mathcal{X}}^{\sigma \triangleleft \tau}$, respectively. \enquote{More abstract} here means a coarse-grained but interventionally consistent representation, telling us essentially the same story about (subjective) cause-and-effect, but at different levels of detail.
    In sheaf jargon, $\rho$ and $\sigma$ are connected by an \textcolor{backbone}{edge $\tau$}, the backbone space is an \textcolor{backbone}{\textit{edge stalk}}, and the mappings are the \textit{restriction maps}.  Therefore, a collection of more abstract representations \textcolor{backbone}{$\{\alpha_{\mathcal{X}}^{\rho \triangleleft \tau}\left(\chi^\rho\right), \alpha_{\mathcal{X}}^{\sigma \triangleleft \tau}\left(\chi^\sigma\right)\}$} is a \textcolor{backbone}{\textit{1-cochain}}. It is now clear that, in our relative setting, global traits of the underlying system only emerge by studying network-level \textit{invariants}. Among them, \enquote{local agreement} of the subjects is particularly important. In sheaf jargon, local agreement refers to \textit{global sections}, i.e., \textcolor{cochain}{0-cochains} whose values are mapped, per each edge, to the same more abstract value-- \textcolor{backbone}{$\alpha_{\mathcal{X}}^{\rho \triangleleft \tau}\left(\chi^\rho\right)= \alpha_{\mathcal{X}}^{\sigma \triangleleft \tau}\left(\chi^\sigma\right)$}. As we show in the bottom right figure, a probability measure \textcolor{cochain}{$\chi^{\rho}$} $\in$ \textcolor{ck}{$\mathsf{CK}(\mathsf{M}^{\rho})$} in the CK of a subject $\rho$ can be mapped to a (usually less informative) probability measure \textcolor{cochain}{$\chi^{\rho,\sigma}$} $\in$ \textcolor{ck}{$\mathsf{CK}(\mathsf{M}^{\rho})$} of another subject $\sigma$, connected to $\rho$ through an edge \textcolor{backbone}{$\tau$}, by first applying a restriction map $\alpha_{\mathcal{X}}^{\rho \triangleleft \tau}$ and then an  extension map $\beta_{\mathcal{X}}^{\sigma \triangleleft \tau}$, i.e., \textcolor{cochain}{$\chi^{\rho,\sigma}=\beta_{\mathcal{X}}^{\sigma \triangleleft \tau}(\alpha_{\mathcal{X}}^{\rho \triangleleft \tau}(\chi^{\rho}))$}. The \textcolor{ck}{\textit{Relative Causal Knowledge} $\mathsf{CK}^{\rho,\sigma}$} of a subject $\rho$ from the perspective of $\sigma$ is then the subspace \textcolor{ck}{$\mathsf{CK}^{\rho,\sigma}$} $\subseteq$ 
 \textcolor{ck}{$\mathsf{CK}(\mathsf{M}^{\sigma})$} being the image of $\beta_{\mathcal{X}}^{\sigma \triangleleft \tau}(\alpha_{\mathcal{X}}^{\rho \triangleleft \tau}(\cdot))$. In the general case, RCK is definable for any pair of subjects for which there exists a connecting path in the underlying network, not only for subjects directly connected by an edge (see \Cref{def:rel_caus_know}).}
    \label{fig:explanation}
\end{figure*}

\noindent\textbf{A Paradigm Shift.} Our work provides a \textit{novel conceptual foundation} for interpreting causality. 
Each subject infers causal knowledge from personal experience and can only see the world through its own perspective. Although the subjectiveness of causality has already been partially explored by \citet{richens2024robust} and, very recently, \citet{bookelias2025causalai}, we move further by making causality not only \textit{subjective} but also \textit{relational}. Consequently, asking for a unique \enquote{true} causal description of a system is an ill-posed question, as the very same notion of causality is inherently relative in our framework. However, it is important to highlight that the relativity of causality is different from the relativism (in its philosophical meaning) of causality: we do not undermine the meaning of things, we question its description as a monolithic object. In our setting, global causal traits of the system emerge only when there is local agreement throughout the network–the global sections of the sheaf of causal knowledge. This parallels how certain physical quantities in physics remain invariant across all reference frames, while other quantities remain inherently relative. But still, the relative treatment better fits the notion of physical reality we are able to describe.  Consider a sudden drop in a company’s stock price. Different agents (the subjects) in the financial system (the network sheaf) propose distinct explanations (their CK): equity analysts blame shifting market sentiment; management highlights an unfavorable earnings report; institutional investors worry about looming regulatory risks; retail shareholders point to negative social media coverage fueling panic selling. However, viewed within the broader network, the interplay of these perspectives (the RCK) discloses a nuanced mix of market sentiment, regulatory uncertainty, company-specific fundamentals, and social influence—no single narrative fully explains the drop. A unified description emerges only when there is local agreement across the entire network (the global sections of the sheaf of causal knowledge). See Figure \ref{fig:explanation} for a simple but comprehensive description of our framework.

\noindent\textbf{The Role of Causal Abstractions.} A subject cannot simply share all of its CK with some other subject because that would lead to an inherently inconsistent notion of perspective and, thus, of relativity. Think about physics: if we move from one reference frame to another, we don't just use the same measurements, but we transform them to make them consistent in the new frame. The notion of perspective is crucial in our framework because it is useful to model a variety of possible elements in a network of subjects: privacy or fairness constraints, maximizing mutual information, or \enquote{simply} modeling the impossibility of a human being to analyze a system if not through their own eyes. In practice, this translates into the uselessness of communicating CK whose distributions have different support on different random variables w.r.t. to the causal knowledge of the receiver subject, which would not know how to use it (see the toy example below). Therefore, the abstractions are a convenient yet rigorous way to model a \enquote{shared discourse space} among subjects to enable communication. In this sense, an abstraction is a backbone space: if the subjects operate on different random variables, abstraction and interventional consistency are arguably the best ways to provide them with a rigorous communication medium.

\noindent\textbf{A Clarification on the Meaning of Cochains.} A network comprises three subjects $\sigma$, $\rho$, and $\gamma$, each of them injecting into the network three soft-intervened probability measures from their own CK. These measures represent their current causal representation of the world. As stated in the main body and in Figure \ref{fig:explanation}, the collection of these three probability measures forms a 0-cochain, i.e., a valuation of the node stalks of the network sheaf. In other words, the value of the 0-cochain at each node is just an object (a probability measure) of the node stalk (the node's CK). Via the restriction maps, the 0-cochain entails a 1-cochain on the edges, i.e., the collection of the abstracted probability measures. Although the subjects share a common CA, the abstracted probability measures might disagree on the edge stalk due to their different understanding of the world at that moment. Then, when a subject, say $\rho$, embeds onto its own node stalk, the abstract probability measure of subjects $\sigma$ and $\gamma$ from the edge stalks, it individuates different probability measures belonging to its CK. The latter represents the understanding of the world of $\sigma$ and $\gamma$ expressed in terms of the causal variables of subject $\rho$. Subject $\rho$ can combine its understanding with those of $\sigma$ and $\gamma$ to accomplish its task, leveraging the closure over convex combinations of the CK (\Cref{th:convex_comb_prob_meas}). A global section can then be seen as a 0-cochain whose values, for each pair of nodes connected by an edge, are mapped to the same 1-cochain value by the restriction maps.

\noindent\textbf{A Toy Example.} Suppose there is some underlying system and consider two subjects $\rho$ and $\sigma$.  The former observes some random variables $X, Z$, and $Y$, and knows that these random variables are related by a structural causal model (SCM) with DAG $X \rightarrow Z \rightarrow Y$. In particular, $\rho$ knows the joint distribution of these variables and also their joint distribution under every possible intervention--that is, $\rho$'s CK. This SCM is compatible with a more abstract SCM whose random variables are $U$ and $M$, and that has DAG $U \rightarrow M$. \enquote{More abstract} here means that there exist surjective mappings porting $X, Z$ and $Y$ into $U$ and $M$. The more abstract SCM is interventionally consistent, i.e., if we pick a variable in the initial SCM, set it to a specific value (\textit{an intervention}), and do the corresponding intervention in the more abstract model, they both tell us essentially the same story about cause-and-effect, but at different levels of detail. Now, suppose $\sigma$ observes variables $P, W, T, L$ and knows they are related by an SCM with DAG $P \rightarrow W \rightarrow T \rightarrow L$. This second SCM is also compatible with the same more abstract SCM with DAG $U \rightarrow M$. Additionally, $\sigma$ can access $\rho$'s CK only through this shared abstract representation and vice versa. If $\sigma$ and $\rho$ are connected by an edge $\tau$, then the Relative CK (RCK) of $\rho$ from the perspective of $\sigma$ is the result of first applying the restriction map to port the CK of $\rho$ on $\tau$, and then porting that more abstract realization of the CK of $\rho$ to $\sigma$ by applying the extension map. In the general case, RCK is definable for any pair of subjects for which there exists a connecting path in the underlying network, not only for subjects directly connected by an edge. Moreover, if more than one possible path connects two subjects, the RCK is path-dependent (see \Cref{def:rel_caus_know}).

\noindent\textbf{Is RCK Needed?} As a little mental exercise to highlight the need for our relative treatment of causality, let us forget for a moment our framework and suppose that there exists a protocol for communicating the entire CK from $\rho$ to $\sigma$ and vice versa. In this case, $\sigma$ receives a certain state (\textit{a cochain value}) of the CK over $\{X, Y, Z\}$ from $\rho$ but, in order to utilize this information, $\sigma$ needs to translate this information onto $\{P, W, T, L\}$. As such, any protocol needs a translation step. The need for this translation step is an informal necessary condition for proving the need of a relative description of causality, beyond our personal philosophical perspective. Consequently, any protocol that uses causal abstraction as the translation step would be a specific instance of our framework. This statement holds true even in more nuanced corner cases: consider again $\rho$ and $\sigma$ but now suppose that both share the same causal model over the same set of random variables, say $\{X, Y, Z\}$. Consider again the protocol above. This time $\sigma$ receives from $\rho$ a certain state of its own causal model according to the perspective of $\rho$ in that context. Then, $\sigma$ can directly combine the $\rho$ 's perspective with its own and accomplish the assigned task. This situation is allowed within our proposed framework. In particular, it is equivalent to set \emph{(i)} the causal abstraction as a causal model over three random variables $\{U, V, W\}$ entailing rotated versions of the probability measures of both $\rho$ and $\sigma$, \emph{(ii)} the restriction maps as rotation matrices, and \emph{(iii)} the extension maps as the transpose of such rotation matrices. The trivial example is when the restrictions and extensions are rotations with a null angle, i.e., the identity.

\noindent\textbf{A Practical and Brief Example in Decentralized Agentic AI.}
In the context of AI, think about the latent spaces of two  autonomous AI agents: they cannot communicate their internal current causal representations (the current 0-cochain) without aligning the latent spaces, i.e., without mapping one latent space into the other. Assume that the latent space of the first agent is a realization of a convex space of probability measures over $\mathbb{R}^d$ and its current causal representation is a point in this space (its current 0-cochain value), and that the latent space of the second agent is a realization of a convex space of probability measures over $\mathbb{R}^n$, with $n \neq d$, and its current causal representation is a point in this space (its current 0-cochain value). The subjects clearly cannot compare or exchange their current causal representations without any additional alignment step.
Consider, however, the simplest scenario: linear alignment, where both restriction and extension maps are represented as matrices.
In this case, the alignment matrix can be viewed as the product of a restriction matrix and an extension matrix. Notably, this factorization aligns closely with the approach commonly employed in low-rank adaptation methods (e.g., LoRA \cite{hu2022lora}). Please notice that this is the case even if the agents communicate in natural language, as they would need to be able to explain to each other how to relate their internal causal representations. At this point, it should be clear as well that the internal current causal representation of one agent can be entirely transmitted to the other agent without any transformation only if the two latent spaces coincide. As a consequence, please notice that the restriction and extension maps are not only useful to allow subjects to communicate, but they can also encode how they should communicate (the perspective). This is why we say that they can be designed to enforce privacy, fairness, or whatever other criterion of interest.